\documentclass[runningheads]{llncs}

\usepackage[utf8]{inputenc} 
\usepackage[T1]{fontenc}    
\usepackage{hyperref}       
\usepackage{url}            
\usepackage{booktabs}       
\usepackage{makecell}       
\usepackage{amsfonts}       
\usepackage{nicefrac}       
\usepackage{microtype}      
\usepackage{adjustbox}

\usepackage{amsmath}
\usepackage{amssymb,bm}
\usepackage{mathtools}

\let\subparagraph\llncssubparagraph

\usepackage{amsthm}
\usepackage{thm-restate}

\usepackage[capitalize,noabbrev,nameinlink]{cleveref}

\usepackage[]{caption}
\usepackage{subcaption}
\usepackage{float}
\usepackage{algorithm,algpseudocode}
\usepackage{latexsym,dsfont}
\usepackage[dvipsnames]{xcolor}
\usepackage{graphicx,graphics,url}
\usepackage{verbatim,ifthen,alltt,array}
\usepackage{multirow}
\usepackage{enumitem}
\usepackage{titlesec}
\usepackage{xparse}
\usepackage{xspace,setspace}

\usepackage{xfrac}
\usepackage[framemethod=TikZ]{mdframed}
\usepackage{nicefrac}       
\usepackage{fontawesome5} 

\usepackage{stmaryrd}
\usepackage{textcomp}
\usepackage{siunitx}
\usepackage{xpatch}
\usepackage{xspace}

\usepackage{pgf}
\usepackage{tikz}
\usepackage{forest}

\usetikzlibrary{arrows,decorations.pathmorphing,decorations.pathreplacing,decorations.footprints,fadings,calc,trees,mindmap,shadows,decorations.text,patterns,positioning,shapes,matrix,fit}
\usetikzlibrary{arrows,shadows,backgrounds}
\usetikzlibrary{arrows.meta}
\usetikzlibrary{positioning,fit}
\usetikzlibrary{automata}
\usetikzlibrary{matrix}
\usetikzlibrary{shapes.symbols,shapes.misc,shapes.arrows}
\usetikzlibrary{matrix,chains,scopes,decorations.pathmorphing}
\usetikzlibrary{bayesnet}
\usetikzlibrary{tikzmark}

\usepackage{tcolorbox}
\usepackage{breakcites}

\usepackage[textsize=tiny]{todonotes}

\definecolor{darkred}{rgb}{0.7,0.1,0.1}
\definecolor{medred}{rgb}{0.5,0.1,0.1}
\definecolor{midred}{rgb}{0.7,0.2,0.2}
\definecolor{vdarkred}{rgb}{0.4,0.1,0.1}
\definecolor{darkslategray}{rgb}{0.18, 0.31, 0.31} 
\definecolor{platinum}{rgb}{0.9, 0.89, 0.89} 
\definecolor{gray}{rgb}{.4,.4,.4}
\definecolor{midgrey}{rgb}{0.5,0.5,0.5}
\definecolor{middarkgrey}{rgb}{0.35,0.35,0.35}
\definecolor{darkgrey}{rgb}{0.3,0.3,0.3}
\definecolor{darkred}{rgb}{0.7,0.1,0.1}
\definecolor{midblue}{rgb}{0.2,0.2,0.7}
\definecolor{darkblue}{rgb}{0.1,0.1,0.5}
\definecolor{darkgreen}{rgb}{0.1,0.5,0.1}
\definecolor{defseagreen}{cmyk}{0.69,0,0.50,0}
\definecolor{purple3}{RGB}{125,38,205}          

\definecolor{tyellow1}{HTML}{FCE94F}
\definecolor{tyellow2}{HTML}{EDD400}
\definecolor{tyellow3}{HTML}{C4A000}
\definecolor{torange1}{HTML}{FCAF3E}
\definecolor{torange2}{HTML}{F57900}
\definecolor{torange3}{HTML}{C35C00}
\definecolor{tbrown1}{HTML}{E9B96E}
\definecolor{tbrown2}{HTML}{C17D11}
\definecolor{tbrown3}{HTML}{8F5902}
\definecolor{tgreen1}{HTML}{8AE234}
\definecolor{tgreen2}{HTML}{73D216}
\definecolor{tgreen3}{HTML}{4E9A06}
\definecolor{tblue1}{HTML}{729FCF}
\definecolor{tblue2}{HTML}{3465A4}
\definecolor{tblue3}{HTML}{204A87}
\definecolor{tpurple1}{HTML}{AD7FA8}
\definecolor{tpurple2}{HTML}{75507B}
\definecolor{tpurple3}{HTML}{5C3566}
\definecolor{tred1}{HTML}{EF2929}
\definecolor{tred2}{HTML}{CC0000}
\definecolor{tred3}{HTML}{A40000}
\definecolor{tlgray1}{HTML}{EEEEEC}
\definecolor{tlgray2}{HTML}{D3D7CF}
\definecolor{tlgray3}{HTML}{BABDB6}
\definecolor{tdgray1}{HTML}{888A85}
\definecolor{tdgray2}{HTML}{555753}
\definecolor{tdgray3}{HTML}{2E3436}

\newtheoremstyle{nthmstyle}
{3pt}
{3pt}
{}
{}
{\bfseries}
{.}
{.5em}
{}

\crefname{enumi}{}{}
\crefname{rstprop}{Proposition}{Propositions}

\newcommand{\fml}[1]{{\mathcal{#1}}}

\newcommand{\msf}[1]{\ensuremath\mathsf{#1}}
\newcommand{\mbf}[1]{\ensuremath\mathbf{#1}}
\newcommand{\mbb}[1]{\ensuremath\mathbb{#1}}

\newcommand{\waxp}{\ensuremath\mathsf{WAXp}}
\newcommand{\wcxp}{\ensuremath\mathsf{WCXp}}
\newcommand{\axp}{\ensuremath\mathsf{AXp}}
\newcommand{\cxp}{\ensuremath\mathsf{CXp}}

\newcommand{\findaxp}{\ensuremath\mathsf{FindAXp}}
\newcommand{\findcxp}{\ensuremath\mathsf{FindCXp}}
\newcommand{\iswaxp}{\ensuremath\mathsf{IsWAXp}}
\newcommand{\iswcxp}{\ensuremath\mathsf{IsWCXp}}
\newcommand{\prooftrace}{\ensuremath\mathsf{ProofTrace}}
\newcommand{\witness}{\ensuremath\mathsf{Witness}}
\newcommand{\witok}{\ensuremath\mathsf{Wit}}

\definecolor{gray}{rgb}{.4,.4,.4}
\definecolor{midgrey}{rgb}{0.5,0.5,0.5}
\definecolor{middarkgrey}{rgb}{0.35,0.35,0.35}
\definecolor{darkgrey}{rgb}{0.3,0.3,0.3}
\definecolor{darkred}{rgb}{0.7,0.1,0.1}
\definecolor{midblue}{rgb}{0.2,0.2,0.7}
\definecolor{darkblue}{rgb}{0.1,0.1,0.5}
\definecolor{defseagreen}{cmyk}{0.69,0,0.50,0}

\newcommand{\jnoteF}[1]{}

\newcounter{Comment}[Comment]
\DeclareMathOperator*{\limply}{\rightarrow}

\declaretheoremstyle[
  headfont=\bfseries,
  bodyfont=\itshape,
  numberwithin=section,
]{StdThmStyle}

\tikzset{
  0 my edge/.style={densely dashed, my edge},
  my edge/.style={-{Stealth[]}},
}

\hypersetup{
  colorlinks = true,
  linkcolor = MidnightBlue,
  urlcolor  = MidnightBlue,
  citecolor = MidnightBlue,
  anchorcolor = MidnightBlue
}

\titleformat{\paragraph}[runin]
{\normalfont\normalsize\bfseries}{\theparagraph}{1em}{}

\newlength{\Oldarrayrulewidth}

\setlist{nosep}

\title{Uncovering Bugs in Formal Explainers}

\subtitle{A Case Study with PyXAI}

\iftrue

\author{%
  Xuanxiang Huang\inst{1} \href{mailto:xuanxiang.huang@ntu.edu.sg}{\small\faEnvelope} \and
  Yacine Izza\inst{2} \and
  Alexey Ignatiev\inst{3} \and
  Joao Marques-Silva\inst{4}
}

\authorrunning{Huang et al.}

\institute{%
  Nanyang Technological University, Singapore
  \and
  CREATE \& National University of Singapore, Singapore
  \and
  Monash University, Australia
  \and
  ICREA \& University of Lleida, Spain
}

\fi

\begin{document}

\maketitle

\begin{abstract}
  Formal explainable artificial intelligence (XAI) offers unique
  theoretical guarantees of rigor when compared to other non-formal
  methods of explainability. However, little attention has been given
  to the validation of practical implementations of formal explainers.
  This paper develops a novel methodology for validating formal
  explainers and reports on the assessment of the publicly available
  formal explainer PyXAI.
  The paper documents the existence of incorrect explanations computed
  by PyXAI on most of the datasets analyzed in the experiments,
  thereby confirming the importance of the proposed novel methodology
  for the validation of formal explainers.
  %
%
\keywords{Formal Explainability \and Validation of Explanations \and
  Proof Tracing \& Checking}
\end{abstract}

\section{Introduction} \label{sec:intro}


Explainable artificial intelligence (XAI) is widely accepted as a key
instrument for delivering trustable
AI~\cite{gunning-darpa19,gunning-sr19,seshia-cacm22}, being critical
for understanding the operation of complex machine learning (ML)
models.
The best-known solutions of XAI are not based on formal
methods~\cite{guestrin-kdd16,guestrin-aaai18,lundberg-nips17}.
These solutions of XAI find an ever-increasing range of uses, but are
also known to lack rigor and often produce erroneous
results~\cite{ignatiev-ijcai20,ms-iceccs23,ms-isola24,msh-cacm24}.
Formal explainability~\cite{darwiche-ijcai18,ms-rw22}, i.e.\ the use
of formal methods for computing rigorous explanations to the
predictions made by ML models, represents an alternative to non-formal
XAI solutions.
One key guarantee of formal XAI are rigorous explanations. Formal 
explanations are most often model-based, and so are guaranteed to be
correct given the ML model.
Recent years witnessed rapid progress in formal
XAI~\cite{ms-rw22,darwiche-lics23,ms-isola24}. At present, scalability
of formal XAI tracks that of adversarial
robustness~\cite{barrett-nips23,ihmpims-kr24}.

Nevertheless, despite a solid theoretical foundation, and like any
other software, practical implementations of formal explainers can 
exhibit bugs.
It is also plain that formal explainers that have bugs will hardly
contribute to building trust in systems of AI, and will undermine the
promised rigor of formal XAI.
%
Furthermore, and to the best of our knowledge, there exists only
initial work on validating formal explainers~\cite{hms-tap23}, but
also on validating verifiers of ML models~\cite{katz-tap25}.
In the case of validating formal explainers, earlier work
\cite{hms-tap23} targets the certification of a formal explanation
algorithm for the concrete case of monotonic classifiers, the
resulting certified explainer is significantly slower than the
original implementation, and it does not address the problem of
validating other families of classifiers.
Moreover, it is believed~\cite{hms-tap23} that the certification of
more complex ML models will be rather more challenging.
%
The purpose of our work is different, as we do not target
certification, but consider instead the practical validation of the
results obtained with a formal explainer.

\paragraph{Contributions.}
This paper develops a framework for validating the explanations
computed by formal explainers. The framework is based on comparing the
outcomes of different formal explainers, but making no assumptions
about the correctness of any of those explainers. One formal explainer
$T$ is the untrusted explainer being validated, i.e.\ the target. A
second formal explainer $R$ is also untrusted, i.e.\ the reference,
but it produces information that aids in validating the obtained
results. In addition, the input of an additional, also untrusted,
formal explainer $S$, i.e.\ the second, is considered in cases where
the information provided by the two explainers $T$ and $R$ does not
suffice to reach a conclusion.

Furthermore, the paper details how the proposed framework has been
used to uncover several bugs in the recently  proposed
explainer for tree-based ML models PyXAI~\cite{marquis-ijcai24a},
i.e.\ the target formal explainer $T$.%
\footnote{%
PyXAI aggregates results from several other publications by some of
the developers of
PyXAI, e.g.~\cite{marquis-kr21,marquis-dke22,marquis-aaai22,marquis-ijcai22b,marquis-aistats23,marquis-ecai23,marquis-ijcai23a}.} 
For that, we used two independently developed formal explainers
RFxpl~\cite{ims-ijcai21} (as the reference formal explainer $R$) and 
XReason~\cite{iisms-aaai22} (as the second formal explainer $S$).
The experiments focus on analyzing the explanations of random
forests~\cite{breiman-ml01}, since for these alternative formal
explainers exist.%
\footnote{XReason~\cite{iisms-aaai22} was developed for boosted trees,
but has been extended with an encoding for random forests for the work
presented in this paper.}

\paragraph{Organization.}
The paper is organized as follows. \cref{sec:prelim} introduces
standard notation and definitions common in machine learning and in
formal explainability.
\cref{sec:approach} details the approach proposed in this paper.
\cref{sec:res} summarizes the results of assessing
PyXAI~\cite{marquis-ijcai24a}.
\cref{sec:conc} concludes the paper.

\jnoteF{Non-trusted non-modifiable non-proof-producing explainer
  vs.\ Non-trusted modifiable proof-producing explainer}


\section{Preliminaries} \label{sec:prelim}
\paragraph{Classification problems.}
%
For simplicity, this paper only considers classification
problems. However, all the concepts discussed in this paper can be
extended to regression problems, as outlined in recent
work~\cite{ms-isola24}.

A classifier is defined on a set of features
$\fml{F}=\{1,\ldots,m\}$. Each feature $i\in\fml{F}$ takes values from
a domain $\mbb{D}_i$, which can be categorical or ordinal and, if
ordinal, can be real- or integer-valued.
Given some fixed order of the features,
e.g.~$\langle1,\ldots,m\rangle$, feature space is defined as the
cartesian product of the domains of the features,
i.e.\ $\mbb{F}=\mbb{D}_1\times\cdots\times\mbb{D}_m$.
A classifier predicts elements from a set $\mbb{K}$ of classes.
Thus, a classifier implements a function
$\kappa:\mbb{F}\to\mbb{K}$. For technical reasons, a classifier is
required not to be constant; however, a constant classifier would be
rather uninteresting.
A classifier is a tuple $M=(\fml{F},\mbb{F},\mbb{K},\kappa)$.
When finding explanations we consider an instance $I=(\mbf{v},c)$,
with $\mbf{v}\in\mbb{F}$, $c\in\mbb{K}$ and $\kappa(\mbf{v})=c$.
Finally, an explanation problem is a tuple $\fml{E}=(M,I)$.

\paragraph{Random forests (RFs).}
%
Similar to other tree ensemble ML models, random forests
(RFs)~\cite{breiman-ml01} were proposed to address the problem of
decision tree overfitting.
A random forest is a set of decision trees, where for each decision
tree the features are picked randomly. Given some input, each tree
picks a class. The model's prediction is then either decided by
majority voting~\cite{breiman-ml01}, or by weighted
voting~\cite{scikitlearn}.
Similar to other tree ensemble models, random forests are well-known
for the quality of their results~\cite{zhou-bk12,zhou-bk21}.
\cref{ssec:cstudies} analyzes the operation of random forests when
assessing the explanations computed by PyXAI~\cite{marquis-ijcai24a}.
A well-known alternative to random forests are boosted
trees~\cite{guestrin-kdd16b}.


\paragraph{Abductive \& contrastive explanations.}
%
We adopt the notation used in earlier
works~\cite{ms-rw22,katz-tacas23,cms-aij23}.
Given an explanation problem $\fml{E}=(M,I)$, with
$M=(\fml{F},\mbb{F},\mbb{K},\kappa)$ and $I=(\mbf{v},c)$, a weak
abductive explanation (WAXp) is a set of features
$\fml{X}\subseteq\fml{F}$ which, if fixed to the values dictated by
$\mbf{v}$, are sufficient to guarantee that the ML model's prediction
is $c$. Formally, we define a predicate
$\waxp:2^{\fml{F}}\to\{\bot,\top\}$:%
\footnote{%
For simplicity, parameterizations are used in the definition of
predicates, being placed as argument after ';'.}
\begin{equation} \label{eq:waxp}
  \waxp(\fml{X};M,I) ~ \coloneq ~
  \forall(\mbf{x}\in\mbb{F}).%
  \left(\bigwedge\nolimits_{i\in\fml{X}}(x_i=v_i)\right)
  \limply\left(\kappa(\mbf{x})=c\right)
\end{equation}
Clearly, the predicate $\waxp$ is monotonically increasing, with
$\waxp(\emptyset)=\bot$ and $\waxp(\fml{F})=\top$.
An abductive explanation (AXp) is a subset-minimal WAXp. Given the
monotonicity of $\waxp$, we define a predicate
$\axp:2^{\fml{F}}\to\{\bot,\top\}$:
\begin{equation} \label{eq:axp}
  \axp(\fml{X};M,I) ~ \coloneq ~
  \waxp(\fml{X};M,I)\land\forall(t\in\fml{X}).%
  \neg\waxp(\fml{X}\setminus\{t\};M,I)
\end{equation}

We can define contrastive explanations in a  similar fashion to
abductive explanations.
Given an explanation problem $\fml{E}=(M,I)$, a weak contrastive
explanation (WCXp) is a set of features $\fml{Y}\subseteq\fml{F}$
which, if the remaining features are fixed to the values dictated by
$\mbf{v}$, then a change of prediction can be attained by changing the
values of the features in $\fml{Y}$. Formally, we define a predicate
$\wcxp:2^{\fml{F}}\to\{\bot,\top\}$: 
\begin{equation} \label{eq:wcxp}
  \wcxp(\fml{X};M,I) ~ \coloneq ~
  \exists(\mbf{x}\in\mbb{F}).%
  \left(\bigwedge\nolimits_{i\in\fml{F}\setminus\fml{Y}}(x_i=v_i)\right)%
  \land\left(\kappa(\mbf{x})\not=c\right)
\end{equation}
Clearly, the predicate $\wcxp$ is monotonically increasing, with
$\wcxp(\emptyset)=\bot$ and $\wcxp(\fml{F})=\top$.
A contrastive explanation (CXp) is a subset-minimal WCXp. Given the
monotonicity of $\wcxp$, we define a predicate
$\cxp:2^{\fml{F}}\to\{\bot,\top\}$:
\begin{equation} \label{eq:cxp}
  \cxp(\fml{X};M,I) ~ \coloneq ~
  \wcxp(\fml{X};M,I)\land\forall(t\in\fml{X}).%
  \neg\wcxp(\fml{X}\setminus\{t\};M,I)
\end{equation}

Furthermore, it is known that $\fml{X}\subseteq\fml{F}$ is an AXp iff
it is a minimal hitting set of the set of CXps, and
vice-versa~\cite{inams-aiia20}.

\paragraph{Connections with automated reasoning.}
By negating~\eqref{eq:waxp} (in the case of WAXps) or directly
from~\eqref{eq:wcxp} (in the case of WCXps), checking whether a set of
features is a WAXp or WCXp can be reduced to deciding the consistency
of a logic formula:
\begin{equation} \label{eq:chksat}
  \bigwedge\nolimits_{i\in\fml{Z}}(x_i=v_i)\land(\kappa(\mbf{x})\not=c)
\end{equation}
Where $\fml{Z}$ is some target set of features and, if consistency
holds, then $\mbf{x}$ represents the point in feature space where the
prediction is no longer $c$, and that is obtained by \emph{not} fixing
the features not in $\fml{Z}$.
Throughout this paper, we will refer to each concrete point $\mbf{w}$
for which~\eqref{eq:chksat} holds and the resulting prediction
$q\in\mbb{K}$ as a \emph{witness}: $(\mbf{w},q)$.
Logic encodings have been devised for different families of
classifiers, which enabled the computation of logic-based explanations
for a wide range of ML models~\cite{ms-rw22,ms-isola24}.
For example, for different families of ML models, \eqref{eq:chksat}
has been decided with SAT \&
MaxSAT~\cite{ims-ijcai21,ims-sat21,iisms-aaai22},
SMT~\cite{inms-aaai19} and MILP~\cite{inms-aaai19}, among others.
Furthermore, and as detailed in earlier
work~\cite{inms-aaai19,ms-rw22}, computing one AXp can be formulated
as the problem of finding one MUS (minimal unsatisfiable subset) of an
inconsistent logic formula. In a similar fashion, computing one CXp
can be formulated as the problem of finding one MCS (minimal
correction subset) on an inconsistent logic formula.
MUSes/MCSes and related concepts are well-established when reasoning
about inconsistent logic formulas~\cite{msm-ijcai20,sat-handbook21}.

%
%
%

\paragraph{Progress in formal XAI.}
%
Since its inception in 2018/19~\cite{darwiche-ijcai18,inms-aaai19},
there has been significant progress in formal XAI, by a growing number
of researchers,
e.g.~\cite{darwiche-ijcai18,inms-aaai19,inms-nips19,darwiche-ecai20,msgcin-nips20,barcelo-nips20,inams-aiia20,kutyniok-jair21,msgcin-icml21,ims-sat21,ims-ijcai21,kwiatkowska-ijcai21a,hiims-kr21,marquis-kr21,cms-cp21,lorini-clar21,mazure-cikm21,marquis-dke22,iisms-aaai22,hiicams-aaai22,marquis-aaai22,rubin-aaai22,darwiche-aaai22,lorini-wollic22,marquis-ijcai22b,amgoud-ijcai22,barcelo-nips22,cms-aij23,ihincms-ijar23,darwiche-jlli23,lorini-jlc23,msi-frai23,yisnms-aaai23,hims-aaai23,marquis-aistats23,hcmpms-tacas23,katz-tacas23,marquis-ijcai23a,hms-ecai23,marquis-ecai23,darwiche-jelia23,ccms-kr23,barrett-nips23,iisms-aaai24,katz-icml24,imms-ecai24,katz-ecai24,ihmpims-kr24,lhms-aaai25,katz-aistats25,bounia-uai25,msllm-ijcai25,iirmss-ijcai25,barcelo-pods25},
among others.


\section{Validation Approach} \label{sec:approach}

\paragraph{Validation of explanations.}
%
Throughout this paper, we assume a given ML model
$M=(\fml{F},\mbb{F},\mbb{K},\kappa)$ and given instance
$I=(\mbf{v},c)$, such that $c=\kappa(\mbf{v})$, thus defining an
explanation problem $\fml{E}=(M,I)$.
To validate that a computed set $\fml{X}\subseteq\fml{F}$ is an AXp,
we follow the definition of AXp in~\eqref{eq:axp}. Concretely, it must
be the case that: (i) $\fml{X}$ is a WAXp; and (ii) for any
$t\in\fml{X}$, $\fml{X}\setminus\{t\}$ is not a WAXp. (Here we exploit
the monotonicity of the $\waxp$ predicate.)
Similarly, to validate that a computed set $\fml{Y}\subseteq\fml{F}$
is a CXp, we follow the definition of CXp in~\eqref{eq:cxp}.
Thus, it must be the case that: (i) $\fml{Y}$ is a WCXp; and (ii)
for any $t\in\fml{Y}$, $\fml{Y}\setminus\{t\}$ is not a WCXp.
(Here we exploit the monotonicity of the $\wcxp$ predicate.)
Thus, a formal explainer will be declared buggy if the testing of any
of the previous conditions fails.

\paragraph{The validation setup.}
%
We consider the following validation scenario. We are given an
untrusted (target) formal explainer $T$, whose results we want to
validate.
Changes to the code of $T$ are only considered to enable the debugging
of the computed results.
We are also given another formal explainer $R$, the reference
explainer, which we will also not trust. The formal explainer $R$ is
capable of producing proof traces, but it is also capable of answering
a number of queries that $T$ is not.
In addition, for some specific validation situations, we will exploit
a second formal explainer $S$, which again we will not necessarily
trust.
Finally, we will assume the ability to query the ML model $M$ on given
points of feature space.
In the remainder of this section, we detail how the checking of the
conditions for AXp/CXp listed above will allow declaring the existence
of errors in $T$ or $R$.

\paragraph{Requirements for the explainer $\bm{T}$.}
%
Given some instance $(\mbf{v},c)$, the untrusted explainer $T$ accepts
the following three queries:
\begin{itemize}[nosep]
\item $\findaxp_T(\mbf{v},c;M)$: returns a set
  $\fml{X}\subseteq\fml{F}$, declared to be an AXp.
\item $\findcxp_T(\mbf{v},c;M)$: returns a set
  $\fml{Y}\subseteq\fml{F}$, declared to be a CXp.
\item $\witness_T(\fml{X},\fml{K})$: returns a point $\mbf{x}$ in
  feature space such that the prediction is a class from some set
  $\fml{K}$ (i.e.\ a subset of $\mbb{K}$).%
  \footnote{%
  In the concrete case of PyXAI~\cite{marquis-ijcai24a}, the semantics
  of witnesses is unclear.
  This remark is motivated by the existing documentation (see
  \url{https://www.cril.univ-artois.fr/pyxai/}).
  As a result, the witnesses of $T$ will \emph{only} be considered as
  such when these are confirmed by the ML model $M$.}
\end{itemize}


\paragraph{Requirements for the reference explainer $\bm{R}$.}
Explainer $R$ is required to provide a superset of the queries
required of explainer $T$.
Thus, and besides those that explainer $T$ also provides, $R$
implements the following queries:
\begin{itemize}[nosep]
\item $\iswaxp_R(\fml{X};M,I)$: checks whether
  $\fml{X}\subseteq\fml{F}$ is a weak AXp for the instance $I$ and
  model $M$.
\item $\iswcxp_R(\fml{X};M,I)$: checks whether
  $\fml{X}\subseteq\fml{F}$ is a weak CXp for the instance $I$ and
  model $M$.
  %
  %
\item $\prooftrace_R(\fml{X},c)$: returns a resolution proof trace for
  why a prediction other than $c$ cannot be attained when the features
  in $\fml{X}$ take the values dictated by $\mbf{v}$.
  Thus, the proof trace is a justification for why $\fml{X}$ is a
  WAXp.
\end{itemize}
The untrusted explainer $T$ may or may not implement the additional
queries above.
%

\paragraph{A second explainer $\bm{S}$.}
%
As shown below, there are cases where the information obtained with
$T$ and $R$ will not suffice to reach a conclusion regarding
correctness. As a result, we
resort to a second formal explainer $S$ as a mechanism to validate (or
not) the results obtained with $R$.
To increase trust, the logic
encoding used in $S$ will be based on SMT, whereas the encoding used
in $R$ will be based on propositional logic, with the condition that
the \emph{same} ML model $M$ is encoded either in propositional logic
or in SMT.%
\footnote{%
Clearly, if different models were considered, then the sets of
(W)AXps/(W)CXps might differ.}
Thus, if $S$ confirms the results of $R$ whenever needed,
then we have additional guarantees regarding the correctness of both
$R$ and $S$. 
Concretely, $S$ is required to implement two queries:
(i) $\iswaxp_S(\fml{X};M,I)$, which checks whether $\fml{X}$ is a weak
AXp for the instance $I$ and model $M$; and (ii)
$\iswcxp_S(\fml{X};M,I)$, which checks whether $\fml{X}$ is a weak
CXp for the instance $I$ and model $M$.

\subsection{Validation of $R$}

As indicated above, we make no assumptions regarding the correctness
of $R$, $S$ and $T$. Accordingly, we will validate all the results
obtained with $R$, as detailed below.

The execution of $R$ when computing AXps/CXps is validated at each
step. If computing an AXp, we iteratively check whether $\waxp$ holds
for a given set of features $\fml{X}$. (One example of an algorithm
for computing an AXp is the well-known \emph{deletion}-based
algorithm~\cite{ms-rw22}, which we assume.)
Thus, if $\neg\waxp(\fml{X})$ holds, then we access the produced
witness, and validate the witness against the ML model $M$. 
Alternatively, if $\waxp(\fml{X})$ holds, then we check the produced
proof trace, and also check with the independent explainer $S$ whether
$\waxp(\fml{X})$ holds.%
\footnote{%
Observe that, since the source of explanation errors may result from
encoding errors, just confirming that a proof is valid is not enough.}
If computing a CXp, we iteratively check whether $\wcxp$ holds for a
given set of features $\fml{Y}$. Thus, if $\wcxp(\fml{Y})$ holds, then 
we access the produced witness, and validate it against the ML model
$M$.
Alternatively, if $\neg\wcxp(\fml{Y})$ holds, then we check the
produced proof trace, and also check with the independent explainer
$S$ whether $\neg\wcxp(\fml{Y})$ holds.
The approach outlined above for validating the results of $R$ are
summarized in~\cref{tab:ref-cases}.
In each of the previous cases, an error is reported if the validation
step fails.

\begin{table}[t]
  \centering
  \renewcommand{\tabcolsep}{0.5em}
  \renewcommand{\arraystretch}{1.15}
  \begin{tabular}{clp{8.575cm}}
    \toprule[1.2pt]
    Case & Condition & Validation \\
    \toprule[1.2pt]
    Ra1 & $\waxp(\fml{X})$ & Validate proof; validate $\waxp(\fml{X}$)
    with $S$
    \\
    Ra2 & $\neg\waxp(\fml{X})$ & Validate witness $(\mbf{w},q)$ with
    model $M$
    \\
    \midrule
    Rc1 & $\wcxp(\fml{Y})$ & Validate witness $(\mbf{w},q)$ with
    model $M$
    \\
    Rc2 & $\neg\wcxp(\fml{Y})$ &  Validate proof; validate
    $\neg\wcxp(\fml{Y}$) with $S$
    \\
    \bottomrule[1.2pt]
  \end{tabular}
  \medskip
  \caption{Validation of reference explainer $R$. If no witness
    exists, then $S$ is used for independently checking the condition
    that no witness exists. If a witness exists, then the witness is
    confirmed with the ML model $M$.
  } \label{tab:ref-cases}
\end{table}


\subsection{Validation of an AXp computed by $T$}

Let the untrusted explainer $T$ compute an AXp
$\fml{X}\subseteq\fml{F}$.
We briefly outline the steps for validating that $\fml{X}$ is an AXp.
\cref{tab:axp-cases} summarizes the different validation scenarios.

\paragraph{The set $\bm{\fml{X}}$ must be a WAXp.}
%
To validate whether $\fml{X}$ is a WAXp, we use the untrusted
explainer $R$ to decide whether $\fml{X}$ is a WAXp.

One outcome is that $R$ concurs that $\fml{X}$ is a WAXp. In this
case, we have two independent explainers confirming that $\fml{X}$ is
a (W)AXp, and so no error is reported (case A1). Furthermore, it is
optional to further validate the result obtained with explainer $R$,
namely by checking that the proof generated by $R$ is indeed a valid
proof, or by using $S$.

Another outcome is that $R$ disagrees with $T$. In this case, $R$
reports a \emph{witness}, i.e.\ a point $\mbf{w}$ in feature space
such that $\kappa(\mbf{w})=q\not=c$ is claimed by $R$. This witness
$(\mbf{w},q)$ is validated against the ML model $M$. If the ML model
$M$ confirms that $\kappa(\mbf{x})=q\not=c$, then we report that $T$ 
produces an incorrect answer, and so we write that a
$\neg\waxp_{T}(\fml{X})$ error was identified (case A2a).%
\footnote{%
Observe that in this case it is \emph{guaranteed} that $T$ produced an
incorrect result.}
However, if the ML model answers that $\kappa(\mbf{w})=c$ (or simply
that $\kappa(\mbf{w})\not=q$), then $R$ has produced an incorrect
answer, and so we write that a $\neg\witok_R(\fml{X};c)$ case was
identified (case A2b).

\paragraph{A proper subset of $\bm{\fml{X}}$ must not be a WAXp.}
%
If both explainers concur that $\fml{X}$ is a WAXp, then we move to a
second validation phase, to decide whether $\fml{X}$ is subset-minimal
for the predicate $\waxp$. 
Because of monotonicity of the predicate $\waxp$, it suffices to check
that $\waxp(\fml{X}\setminus\{t\})$ does not hold for each
$t\in\fml{X}$.

For each $t\in\fml{X}$, we use the explainer $R$, to decide whether
$\neg\waxp_R(\fml{X}\setminus\{t\})$ holds.
If that is the case, then a witness is obtained as follows: 
$(\mbf{w},q)=\witness_R(\fml{X}\setminus\{t\},\mbb{K}\setminus\{c\})$.
The witness is validated against the ML model $M$. If the model does
not confirm a change in prediction, then this represents an error
situation for the explainer $R$ (case A4b). Otherwise, no error is
reported (case A4a).

If for some $t\in\fml{X}$, it is the case that
$\waxp_R(\fml{X}\setminus\{t\})$ holds, then we use the proof trace
produced by the explainer $R$ to confirm that it is a valid proof.
If the proof is not validated, then we report an error with the
reasoner used by $R$ (case A3b).
If the proof is validated, then we use the second explainer $S$ to
independently check whether $\waxp_S(\fml{X}\setminus\{t\})$ is a
WAXp. If $S$ concurs that $\fml{X}\setminus\{t\}$ is a WAXp, then we
report an error for $T$ (case A3a). Otherwise, we report an error for
$R$ (case A3c).
%
%

\begin{table}[t]
  \centering
  \renewcommand{\tabcolsep}{0.5em}
  \renewcommand{\arraystretch}{1.15}
  \begin{tabular}{cp{5.25cm}p{5.25cm}}
    \toprule[1.2pt]
    Case & Description & Verdict \\
    \toprule[1.2pt]
    A1 & $R$ confirms that $\fml{X}$ is WAXp & No error reported
    \\
    \midrule
    A2a & $R$ reports witness $(\mbf{w},q)$; witness validated by $M$
    & \makecell[tl]{$\neg\waxp$ error for $T$:\\$\fml{X}$ computed by $T$ is not WAXp}
    \\
    \midrule
    A2b & $R$ reports witness $(\mbf{w},q)$; witness \emph{not}
    validated by $M$
    & \makecell[tl]{$R$ error:\\$(\mbf{w},q)$ is not witness}
    \\
    \midrule
    %
    %
    A3a & $R$ reports $\fml{X}\setminus\{t\}$ is WAXp, proof checked
    \& WAXp confirmed by $S$ 
    & \makecell[tl]{$\waxp\land\neg\axp$ error for $T$:\\
      Opting to trust $R$ and proof\\
      as long as $S$ concurs with $R$} 
    \\
    \midrule
    A3b & $R$ reports $\fml{X}\setminus\{t\}$ is WAXp \& proof
    \emph{not} checked & Issue with reasoner used by $R$
    \\
    \midrule
    A3c & $R$ reports $\fml{X}\setminus\{t\}$ is WAXp, proof
    checked \& $S$ does not agree with $R$ &
    \makecell[tl]{$R$ error:\\ $\fml{X}\setminus\{t\}$ not WAXp}
    \\
    \midrule
    A4a & $R$ reports witness $(\mbf{w},q)$; witness validated by $M$
    &
    No error reported
    \\
    \midrule
    A4b & $R$ reports witness $(\mbf{w},q)$; witness \emph{not}
    validated by $M$
    &
    \makecell[tl]{$R$ error:\\$(\mbf{w},q)$ is not witness}
    \\
    \bottomrule[1.2pt]
  \end{tabular}
  \medskip
  \caption{Validating AXp $\fml{X}\subseteq\fml{F}$ computed with $T$,
    using $R$ and (if necessary) $S$.
    If no errors reported for $T$ for AXp $\fml{X}$, then this counts
    as a validated AXp.
  } \label{tab:axp-cases}
\end{table}

\subsection{Validation of a CXp computed by $T$}

Let the untrusted explainer $T$ compute a CXp
$\fml{Y}\subseteq\fml{F}$.
We briefly outline the steps for validating that $\fml{Y}$ is a CXp.
\cref{tab:cxp-cases} summarizes the different validation scenarios.

\paragraph{Set $\bm{\fml{Y}}$ must be a WCXp.}
%
To validate whether $\fml{Y}$ is a WCXp, we check whether the witness
produced by the explainer $T$ is confirmed by the ML model $M$.
Such a witness $\mbf{w}$ must satisfy the following condition: 
$w_i = v_i$ for any feature $i \notin \fml{Y}$.
If it is, then we accept $\fml{Y}$ as a WCXp (case C1a). (This step only
serves to confirm the result of $T$, not to declare errors.)
Otherwise, we proceed to analyze $\fml{Y}$ with the explainer $R$.
%
If $R$ confirms that $\fml{Y}$ is a CXp, then we obtain a witness
$(\mbf{w},q)$ from $R$, which we validate against the classifier $M$.
If the ML model $M$ accepts the witness, then we have independent
confirmation that the CXp reported by $T$ is guaranteed to be a WCXp
(case C1b). 
If the ML model $M$ does not accept the witness, then we report that
$R$ is buggy (case C1c).

If $R$ fails to confirm that $\fml{Y}$ is a WCXp, then the proof trace 
can be checked for correctness of the proof.
If the proof is not validated, then we report an error with the
reasoner used by $R$ (case C2a). Otherwise, we further check that $\fml{Y}$ is
not a WCXp with the second explainer $S$. If $S$ concurs with $R$,
then we report an error with $T$ (case C2b). Otherwise we report an
error with $R$ (case C2c).

\paragraph{A proper subset of $\bm{\fml{Y}}$ must not be a WCXp.}
%
Assuming that $\fml{Y}$ is accepted to be a WCXp, by both by $T$ and
$R$, then we need to check minimality. We check each
$\fml{Y}\setminus\{t\}$, with $t\in\fml{Y}$, to confirm that none is a
WCXp.

If for some $t\in\fml{Y}$, $R$ reports that $\fml{Y}\setminus\{t\}$ to
be a WCXp, then we obtain a witness $(\mbf{w},q)$, with $q\not=c$, to
that effect, and check the witness with the ML model $M$. If $M$
confirms that the witness $(\mbf{w},q)$, then we have an example where
$T$ computes a non-minimal CXp (case C3a).%
\footnote{%
Observe that in this case it is \emph{guaranteed} that $T$ produced an
incorrect result.}
Otherwise, we have an example of a non-witness computed by $R$, and so
$R$ is in error (case C3b).%
\footnote{As indicated earlier, any error regarding $R$ immediately
stops the validation of $T$.}
If $R$ confirms that $\fml{Y}\setminus\{t\}$ is not a WCXp, then the
proof trace is checked. If some error is detected, then an error is
reported regarding the reasoner used by $R$ (case C4b). Otherwise, no
error is reported (case C4b).

\begin{table}[t]
  \centering
  \renewcommand{\tabcolsep}{0.5em}
  \renewcommand{\arraystretch}{1.15}
  \begin{tabular}{cp{5.25cm}p{5.25cm}}
    \toprule[1.2pt]
    Case & Description & Verdict \\
    \toprule[1.2pt]
    C1a & $T$ generates witness $(\mbf{w},q)$ \& $M$ 
    validates witness &
    No error reported
    \\
    \midrule
    C1b & $R$ confirms that $\fml{Y}$ is WCXp with witness
    $(\mbf{w},q)$ \& $M$ validates witness & No error reported
    \\
    \midrule
    C1c & $R$ confirms that $\fml{Y}$ is WCXp with witness
    $(\mbf{w},q)$ \& $M$ does \emph{not} validate witness &
    \makecell[tl]{$R$ error:\\$(\mbf{w},q)$ is not witness}
    \\
    \midrule
    C2a & $R$ reports that $\fml{Y}$ is not WCXp, proof \emph{not}
    validated
    & Issue with reasoner used by $R$
    \\
    \midrule
    C2b & $R$ reports that $\fml{Y}$ is not WCXp, proof validated
    \& $S$ agrees with $R$
    & \makecell[tl]{$\neg\wcxp$ error for $T$:\\
      Opting to trust $R$ and proof\\
      as long as $S$ concurs with $R$
    }
    \\
    \midrule
    C2c & $R$ reports that $\fml{Y}$ is not WCXp, proof validated
    \& $S$ disagrees with $R$
    & \makecell[tl]{$R$ error:\\$\fml{Y}$ computed by $T$ may be WCXp}
    \\
    \midrule
    C3a & $R$ reports $\fml{Y}\setminus\{t\}$ is WCXp \& witness
    validated by $M$
    & \makecell[tl]{$\wcxp\land\neg\cxp$ error for $T$}
    \\
    \midrule
    C3b & $R$ reports $\fml{Y}\setminus\{t\}$ is WCXp \& witness
    \emph{not} validated by $M$
    & \makecell[tl]{$R$ error:\\$(\mbf{w},q)$ is not a witness}
    \\
    \midrule
    C4a & $R$ confirms $\fml{Y}\setminus\{t\}$ not WCXp \& proof
    \emph{not} validated
    &
    \makecell[tl]{Issue with reasoner used by $R$}
    \\
    \midrule
    C4b &$R$ confirms $\fml{Y}\setminus\{t\}$ not WCXp \& proof
    validated 
    & No error reported
    \\
    \bottomrule[1.2pt]
  \end{tabular}
  \medskip
  \caption{Validating CXp $\fml{Y}\subseteq\fml{F}$ computed with $T$,
    using $R$ and (if necessary) $S$.
    If no errors reported for $T$ for CXp $\fml{Y}$, then this counts
    as a validated CXp.
  } \label{tab:cxp-cases}
\end{table}

\subsection{Handling Bugs in $R$}

As stated earlier in this section, the correctness of $R$ is assessed
at every step, when computing explanations, or when validating the
explanations of $T$.
If at any step any error is identified with respect to $R$, then we
opt not to make \emph{any} claims regarding the correctness of $T$,
and simply report $R$ to be buggy. As the results underline, that was
never the case in our experiments.
Nevertheless, future work will contemplate extending the initial ideas
on certification of explainers~\cite{hms-tap23}, to the case of other
families of ML models. Certified explainers, obtained with a proof
assistant like Rocq~\cite{bertot-bk04} or Lean~\cite{moura-cade21},
should be expected to be significantly slower than non-certified
explainers. However, these could be used only as the second (but now
trusted) explainer $S$.


\section{Experiments \& Case Studies} \label{sec:res}

\subsection{Experiments}

\paragraph{Goals.}
%
The experiments evaluate the proposed explanation validation approach,
specifically targeting explanations produced by the well-known and
publicy available explanation toolkit PyXAI~\cite{marquis-ijcai24a}.
With respect to the proposed validation approach, PyXAI will
correspond to explainer $T$, RFxpl~\cite{ims-ijcai21} will correspond
to explainer $R$, and XReason~\cite{iisms-aaai22} will correspond to
the second explainer $S$.%
\footnote{For these experiments, XReason was extended with an SMT
encoding for random forests.}

The first experiment validates PyXAI~\cite{marquis-ijcai24a} when
computing AXps for random forests (RFs). The second experiment
validates PyXAI when computing CXps again for RFs.
In addition, the existing prototype RFxpl has been extended to output
both witnesses and proof traces obtained from the underlying SAT
solver.
Furthermore, in the case of computation of CXps, PyXAI was extended to
output a witness. However, we observed that many of the witnesses produced by
PyXAI are incorrect, i.e.\ the ML model $M$ fails to confirm that the
witnesses cause a change of prediction.
As a result, we opt not to report those results, and only use the
witnesses obtained from PyXAI when RFxpl reports that a CXp computed
by PyXAI is \emph{not} a WCXp. And, PyXAI's witnesses can only serve
to prove that PyXAI's results are correct, and \emph{not} to argue for
PyXAI's incorrectness.

The source code for reproducing our results is available at~\url{https://github.com/XuanxiangHuang/chk-fxpr}.

\paragraph{Experimental setup.}
Experiments for computing explanations with PyXAI were conducted within the Docker environment provided by the PyXAI framework.
Validation experiments for PyXAI's explanation were performed on a MacBook Pro equipped with a 6-core
Intel Core i7 (2.6 GHz) processor and 16 GB RAM, running macOS Sequoia.
The evaluation involved 27 binary classification datasets selected from
ADBench~\cite{han2022adbench},%
\footnote{\url{https://github.com/Minqi824/ADBench/tree/main/adbench/datasets/Classical}}
Penn Machine Learning Benchmarks~\cite{olson2017pmlb},%
\footnote{\url{https://epistasislab.github.io/pmlb/}}
and
UCI Machine Learning
Repository~\cite{asuncion2007uci}.%
\footnote{\url{https://archive.ics.uci.edu/}}
Additional details of the training of the RFs used in the experiments
is included in~\cref{sec:training}.
The explanation tools used in this evaluation include
RFxpl,%
\footnote{\url{https://github.com/izzayacine/RFxpl}}
XReason,%
\footnote{\url{https://github.com/alexeyignatiev/xreason}}
and PyXAI.%
\footnote{\url{https://github.com/crillab/pyxai/tree/6fe3c85b24d13bab71be0a6b5ec478a326495b67};
we used the latest version of PyXAI released on May 6, 2025.}
For the evaluation of explanation tools, we randomly selected 200 instances from each dataset (or all instances if fewer than 200 were available).
For each instance, we computed one AXp and one CXp using PyXAI, and the explanations were then checked by the RFxpl and XReason tools.
Moreover, the proof checker DRAT-trim~\cite{wetzler2014drat}%
\footnote{\url{https://github.com/marijnheule/drat-trim}}
was used to verify the proof traces produced by the SAT solver employed by RFxpl.

\paragraph{Validation of RFxpl.}
In all the experiments, \emph{no} examples were identified indicating
the existence of errors in the case of RFxpl. This includes \emph{all}
the possible issues detailed
in~\cref{tab:ref-cases,tab:axp-cases,tab:cxp-cases}.%
%
%
\footnote{%
In one case, validation using RFxpl failed, but due to the lack of
available memory in the computer used for the experiments.}

Although the results do not prove the absence of errors in RFxpl,
they do increase our trust in the assessment of PyXAI.

\paragraph{Validation of PyXAI's AXps.}
%
%
\cref{tab:rf_pyxai_axp_quality} summarizes the results obtained regarding
AXps computed with PyXAI.
The errors reported in column \%$[\neg\waxp]$ are \emph{guaranteed} to
be the result of bug(s) in PyXAI.
The errors reported in column \%$[\waxp\land\neg\axp]$ represent cases
where PyXAI claims subset-minimality and both RFxpl and XReason concur
that the explanation is not subset-minimal. (In addition, confidence
in these results is supported by the fact that no error was observed in
any of the experiments that use RFxpl and generated witnesses.)

\begin{table}[t]
  \centering
  \renewcommand{\tabcolsep}{0.725em}
\begin{tabular}{lrrr}
\toprule[1.2pt]
Dataset             &  \%$\:[\neg\msf{WAXp}]$ & \%$\:[\msf{WAXp} \land \neg \msf{AXp}] $ & \%$\:[\msf{AXp}]$ \\
\midrule[1.2pt]
12\_fault           & 0.0\%        & 100.0\%           & 0.0\%      \\
21\_Lymphography    & 0.0\%        & 97.3\%            & 2.7\%      \\
29\_Pima            & 0.0\%        & 99.0\%            & 1.0\%      \\
30\_satellite       & 0.0\%        & 100.0\%           & 0.0\%      \\
31\_satimage-2      & 0.0\%        & 100.0\%           & 0.0\%      \\
33\_skin            & 0.0\%        & 37.6\%            & 62.4\%     \\
37\_Stamps          & 0.0\%        & 99.5\%            & 0.5\%      \\
4\_breastw          & 0.0\%        & 98.7\%            & 1.3\%      \\
6\_cardio           & 0.0\%        & 100.0\%           & 0.0\%      \\
7\_Cardiotocography & 0.0\%        & 100.0\%           & 0.0\%      \\
appendicitis        & 0.0\%        & 98.1\%            & 1.9\%      \\
banknote            & 0.0\%        & 100.0\%           & 0.0\%      \\
biodegradation      & 0.0\%        & 100.0\%           & 0.0\%      \\
glass2              & 0.0\%        & 99.4\%            & 0.6\%      \\
heart-c             & 0.0\%        & 99.5\%            & 0.5\%      \\
ionosphere          & 0.0\%        & 100.0\%           & 0.0\%      \\
magic               & 0.0\%        & 100.0\%           & 0.0\%      \\
mofn-3-7-10         & 0.0\%        & 0.0\%             & 100.0\%    \\
phoneme             & 0.0\%        & 97.5\%            & 2.5\%      \\
ring                & 0.0\%        & 97.5\%            & 2.5\%      \\
sonar               & 1.5\%        & 98.0\%            & 0.5\%      \\
spambase            & 0.0\%        & 100.0\%           & 0.0\%      \\
spectf              & 1.0\%        & 99.0\%            & 0.0\%      \\
twonorm             & 0.0\%        & 100.0\%           & 0.0\%      \\
wdbc                & 0.0\%        & 100.0\%           & 0.0\%      \\
wpbc                & 0.0\%        & 100.0\%           & 0.0\%      \\
xd6                 & 2.5\%        & 0.0\%             & 97.5\%    \\
\bottomrule[1.2pt]
\end{tabular}
\medskip
\caption{
    Assessing the quality of AXps for RFs generated by the PyXAI tool
    with default settings (using MUSER Solver).
    \%$[\msf{AXp}]$ denotes the fraction of computed AXps that are
    validated as such (i.e.\ cases A1\&A4a in~\cref{tab:axp-cases}), 
    \%$[\neg\msf{WAXp}]$ denotes the fraction of computed AXps that
    are incorrect, not even representing WAXps (case A2a, and a
    provable bug for PyXAI), and \%$[\msf{WAXp}\land\neg\msf{AXp}]$
    denotes the fraction of computed  AXps that are redundant (case
    A3a, according to both RFxpl and Xreason).
    All RFs have 50 trees. The number of instances is 200, or 
    all instances if there are fewer than 200.
    The time limit was set to 60 seconds; no timeout was observed.
}
\label{tab:rf_pyxai_axp_quality}
\end{table}

\paragraph{Validation of PyXAI's CXps.}
%
\cref{tab:rf_pyxai_cxp_quality} summarizes the results obtained
regarding CXps computed with PyXAI.
The errors reported in columns \%$[\neg\wcxp]$ represent cases where
PyXAI claims a WCXp, when both RFxpl and XReason concur that that is
not the case. (As stated above, confidence in these results is
supported by the fact that no error was observed in any of the
experiments that use RFxpl and generated witnesses.)
The errors reported in column \%$[\wcxp\land\neg\cxp]$ are
\emph{guaranteed} to be the result of bug(s) in PyXAI.
(Observe that the large fraction of guaranteed non-CXps that are
reported as CXps by PyXAI, and the methodology detailed in this paper,
will be expected to help decisively the developers of PyXAI in
locating the existing bug(s).)

\begin{table}[t]
  \centering
  \renewcommand{\tabcolsep}{0.725em}
\begin{tabular}{lrrr}
\toprule[1.2pt]
Dataset             &  \%$\:[\neg\msf{WCXp}]$ & \%$\:[\msf{WCXp} \land \neg \msf{CXp}]$ & \%$\:[\msf{CXp}]$ \\
\midrule[1.2pt]
12\_fault           & 0.0\%        & 83.0\%            & 17.0\%     \\
21\_Lymphography    & 6.1\%        & 56.1\%            & 37.8\%     \\
29\_Pima            & 0.0\%        & 81.0\%            & 19.0\%     \\
30\_satellite       & 0.0\%        & 97.5\%            & 2.5\%      \\
31\_satimage-2      & 0.0\%        & 100.0\%           & 0.0\%      \\
33\_skin            & 0.0\%        & 41.8\%            & 58.2\%     \\
37\_Stamps          & 0.0\%        & 96.0\%            & 4.0\%      \\
4\_breastw          & 0.0\%        & 97.3\%            & 2.7\%      \\
6\_cardio           & 0.0\%        & 97.0\%            & 3.0\%      \\
7\_Cardiotocography & 0.5\%        & 75.5\%            & 24.0\%     \\
appendicitis        & 0.0\%        & 98.1\%            & 1.9\%      \\
banknote            & 0.0\%        & 96.5\%            & 3.5\%      \\
biodegradation      & 0.0\%        & 79.0\%            & 21.0\%     \\
glass2              & 0.0\%        & 82.7\%            & 17.3\%     \\
heart-c             & 0.0\%        & 5.5\%             & 94.5\%     \\
ionosphere          & 0.0\%        & 94.5\%            & 5.5\%      \\
magic               & 0.0\%        & 95.0\%            & 5.0\%      \\
mofn-3-7-10         & 0.0\%        & 4.0\%             & 96.0\%     \\
phoneme             & 0.0\%        & 92.0\%            & 8.0\%      \\
ring                & 0.0\%        & 89.5\%            & 10.5\%     \\
sonar               & 0.0\%        & 95.5\%            & 4.5\%      \\
spambase            & 0.0\%        & 97.0\%            & 3.0\%      \\
spectf              & 0.0\%        & 84.0\%            & 16.0\%     \\
twonorm             & 0.0\%        & 98.5\%            & 1.5\%      \\
wdbc                & 0.0\%        & 98.5\%            & 1.5\%      \\
wpbc                & 2.1\%        & 93.8\%            & 4.1\%      \\
xd6                 & 0.0\%        & 0.0\%             & 100.0\%   \\
\bottomrule[1.2pt]
\end{tabular}
\medskip
\caption{
    Assessing the quality of CXps for RFs obtained using the PyXAI
    tool with default settings (using OPENWBOSolver).
    \%$[\msf{CXp}]$ denotes the fraction of computed CXps that are
    validated as such (i.e.\ cases C1a\&C4a in~\cref{tab:cxp-cases}), 
    \%$[\neg\msf{WCXp}]$ denotes the fraction of computed CXps that
    are incorrect, not even representing WCXps (case C2a, according to
    both RFxpl and Xreason), and \%$[\msf{WCXp}\land\neg\msf{CXp}]$
    denotes the fraction of computed CXps that are redundant (case
    C3a, and a provable bug for PyXAI).
    All RFs have 50 trees. The number of instances is 200, or 
    all instances if there are fewer than 200.
    The time limit was set to 60 seconds; no timeout was observed.
}
\label{tab:rf_pyxai_cxp_quality}
\end{table}

\begin{table}[t]
  \centering
  \renewcommand{\tabcolsep}{0.85em}
  \renewcommand{\arraystretch}{1.15}
  \begin{tabular}{lrrr}
    \toprule[1.2pt]
    & \multicolumn{2}{c}{\# Validation scheme cases}
    &
    \\
    \cmidrule[0.95pt]{2-3}
    \multirow{1}{*}{Type}
    & 
    \multicolumn{1}{c}{Witness~\&~ML~model} &
    \multicolumn{1}{c}{XReason} &
    \multirow{1}{*}{Total}
    \\
    \midrule[1.2pt]
    WAXp & 42,165 & 34,812 & 76,977 \\
    WCXp & 21,800 & 14,984 & 36,784 \\
    \midrule[0.925pt]
    Total & 63,965 & 49,796 & 113,761 \\
    \bottomrule[1.2pt]
  \end{tabular}
  \medskip
  \caption{Distribution of WAXp and WCXp calls according to validation
    scheme. When $\waxp()$ holds or $\wcxp()$ does not hold, the results
    are further validated using XReason. Conversely, when $\waxp()$ does
    not hold or $\wcxp()$ holds, the results are justified by validating 
    the witness produced by RFxpl with the ML model $M$.}
  \label{tab:rf_pyxai_stats2}
\end{table}

\begin{table}[t]
\centering
\begin{tabular}{lr}
\toprule[1.2pt]
Validation scheme & \multicolumn{1}{c}{\# PyXAI errors} \\
\midrule[1.2pt]
Witness \& ML model validation        & 16,661 (non-min.) + 198 (incorrect) \\
RFxpl \& XReason agreeing on result       & 29,669 (non-min.) + 51 (incorrect) \\
\bottomrule[1.2pt]
\end{tabular}
\medskip
\caption{The number of incorrect explanation obtained with PyXAI under
  the two proposed validation schemes. As detailed earlier, an
  incorrect explanation is declared for cases where the 
  explanation is either not valid or not subset-minimal.} 
\label{tab:rf_pyxai_stats3}
\end{table}

\paragraph{Additional statistics.}
%
For the experimental results shown above,
%
\cref{tab:rf_pyxai_stats2} shows the number of $\waxp$ and
$\wcxp$ oracle calls that have been checked with RFxpl, and then
either with the ML model $M$ or with XReason.
Moreover, \cref{tab:rf_pyxai_stats3} dissects the cases where the 
results of PyXAI failed to be validated. 
%
The relatively large number of cases where PyXAI is declared buggy due
to validation with RFxpl and XReason is justified by the relatively
larger fraction of AXps that is subject to validation.

\paragraph{Beyond random forests.}
The approach proposed in this paper to validate a formal explainer has
also been applied to boosted trees~\cite{chen2016xgboost}, with the
existence of bug(s) in PyXAI also observed. \cref{sec:bts} summarizes
these additional experimental results.

\paragraph{Confidence in the results.}
Given the large percentage of guaranteed incorrect explanations
obtained with PyXAI, namely columns \%$[\neg\waxp]$
in~\cref{tab:rf_pyxai_axp_quality} and \%$[\wcxp\land\neg\cxp]$
in~\cref{tab:rf_pyxai_cxp_quality}, and given the fact that no errors
were observed with RFxpl and XReason, we are confident that the
non-guaranteed incorrect explanations observed with PyXAI, namely
columns \%$[\waxp\land\neg\axp]$ in~\cref{tab:rf_pyxai_axp_quality}
and \%$[\neg\wcxp]$ in~\cref{tab:rf_pyxai_cxp_quality}, are also the
result of bug(s) in PyXAI.

\subsection{Case Studies} \label{ssec:cstudies}

This section provides additional detail regarding PyXAI's identified
bug(s). We analyze two very simple RFs, each containing enough detail to 
demonstrate one of PyXAI's bug(s).
The first case study illustrates case A2a (see~\cref{tab:axp-cases}),
and so it is a guranteed bug of PyXAI. The second case study
illustrates case C2b (see~\cref{tab:cxp-cases}). Although this bug
results from both RFxpl and XReason disagreeing with PyXAI, and so one
could argue that the bug's existence is not certain, for the second
case study, the existence of a bug in PyXAI is confirmed. In general,
this additional manual validation step can only be implemented for
very simple models.

\paragraph{Example of $\bm{\neg\waxp}$ error in PyXAI.} 
%
We describe a simple RF for which PyXAI computes an AXp $\fml{X}$,
which is shown not to be a WAXp. This corresponds to case A2a
in~\cref{tab:axp-cases}.

\begin{figure}[t]
\centering
\begin{adjustbox}{center}
\setlength{\tabcolsep}{8pt}
\def\arraystretch{3}
    \begin{tabular}{cc}
        \scalebox{0.725}{\forestset{
  BDT/.style={
    for tree={
      l=1.125cm,s sep=0.575cm,
      if n children=0{}{circle},
      draw,
      edge={
        my edge
      },
      if n=2 {
        edge+={0 my edge},
      }{},
    }
  },
}
\begin{forest}
 tikz+={\node[anchor=south] at (.north) {\text{T\(_1\)}};},
  BDT
  [$x_1$
    [$x_9$,  edge label={node[near start,left,xshift=-2.75pt] {{\scriptsize0}}}
      [{\footnotesize\color{darkgreen}0}]
      [$x_7$
        [{\footnotesize\color{darkred}0}]
        [{\footnotesize\color{darkgreen}1}]
      ]
    ]
    [$x_2$, edge label={node[near start,right,xshift=2.75pt] {{\scriptsize1}}}, edge+={ultra thick, draw=darkgreen}
     [$x_8$, edge+={ultra thick, draw=darkgreen}
       [{\footnotesize\color{darkred}0}]
       [$x_9$, edge+={ultra thick, draw=darkgreen} 
         [{\footnotesize\color{darkred}0}]
         [{\footnotesize\color{darkgreen}1}, edge+={ultra thick, draw=darkgreen}]
       ]  
     ]
     [$x_5$
       [$x_8$ 
         [{\footnotesize\color{darkred}0}]  
         [{\footnotesize\color{darkgreen}1}]
       ]
       [$x_3$
         [{\footnotesize\color{darkred}0}]  
         [{\footnotesize\color{darkgreen}1}]       
       ]
     ]
    ]
  ]
\end{forest}} & \scalebox{0.725}{\forestset{
  BDT/.style={
    for tree={
      l=1.125cm,s sep=0.575cm,
      if n children=0{}{circle},
      draw,
      edge={
        my edge
      },
      if n=2 {
        edge+={0 my edge},
      }{},
    }
  },
}
\begin{forest}
 tikz+={\node[anchor=south] at (.north) {\text{T\(_2\)}};},
  BDT
  [$x_7$
    [$x_2$, edge label={node[near start,left,xshift=-2.75pt] {{\scriptsize0}}}
      [{\footnotesize\color{darkred}0}]
      [$x_6$
        [{\footnotesize\color{darkred}0}]
        [$x_3$
          [{\footnotesize\color{darkred}0}]
          [{\footnotesize\color{darkgreen}1}]
        ]
      ]
    ]  
    [$x_6$, edge label={node[near start,right,xshift=2.75pt] {{\scriptsize1}}}, edge+={ultra thick, draw=darkgreen}
     [$x_9$
      [{\footnotesize\color{darkred}0}]
      [$x_2$
         [{\footnotesize\color{darkred}0}]
         [{\footnotesize\color{darkgreen}1}]
      ] 
     ]
     [$x_5$, edge+={ultra thick, draw=darkgreen}
      [{\footnotesize\color{darkred}0}]
      [{\footnotesize\color{darkgreen}1}, edge+={ultra thick, draw=darkgreen}]
     ] 
    ]
  ]
\end{forest}} \\
         \scalebox{0.725}{\forestset{
  BDT/.style={
    for tree={
      l=1.125cm,s sep=0.575cm,
      if n children=0{}{circle},
      draw,
      edge={
        my edge
      },
      if n=2 {
        edge+={0 my edge},
      }{},
    }
  },
}
\begin{forest}
 tikz+={\node[anchor=south] at (.north) {\text{T\(_3\)}};},
  BDT
  [$x_7$
    [$x_1$, edge label={node[near start,left,xshift=-2.75pt] {{\scriptsize0}}}
      [{\footnotesize\color{darkred}0}]
      [$x_2$
         [{\footnotesize\color{darkred}0}]
         [$x_8$
           [{\footnotesize\color{darkgreen}1}]
           [{\footnotesize\color{darkred}0}]
         ]
      ]
    ]
    [$x_9$, edge label={node[near start,right,xshift=2.75pt] {{\scriptsize1}}}, edge+={ultra thick, draw=darkgreen}
      [$x_5$
         [{\footnotesize\color{darkred}0}]
         [$x_6$
           [{\footnotesize\color{darkred}0}]
           [{\footnotesize\color{darkgreen}1}]
         ]
      ]
     [$x_8$, edge+={ultra thick, draw=darkgreen} 
       [{\footnotesize\color{darkred}0}]
       [{\footnotesize\color{darkgreen}1}, edge+={ultra thick, draw=darkgreen}]
     ]
    ]
  ]
\end{forest}} & \scalebox{0.725}{\forestset{
  BDT/.style={
    for tree={
      l=1.125cm,s sep=0.575cm,
      if n children=0{}{circle},
      draw,
      edge={
        my edge
      },
      if n=2 {
        edge+={0 my edge},
      }{},
    }
  },
}
\begin{forest}
 tikz+={\node[anchor=south] at (.north) {\text{T\(_4\)}};},
  BDT
  [$x_6$
    [$x_9$, edge label={node[near start,left,xshift=2.75pt] {{\scriptsize0}}}
     [{\footnotesize\color{darkred}0}]
     [$x_3$
      [{\footnotesize\color{darkred}0}] 
      [$x_8$
       [{\footnotesize\color{darkred}0}]
       [{\footnotesize\color{darkgreen}1}]
      ]
     ]
    ]
    [$x_5$, edge label={node[near start,right,xshift=2.75pt] {{\scriptsize1}}}, edge+={ultra thick, draw=darkgreen}
     [$x_8$
      [{\footnotesize\color{darkred}0}]
      [$x_4$
       [{\footnotesize\color{darkred}0}]
       [{\footnotesize\color{darkgreen}1}]
      ]
     ]
     [$x_3$, edge+={ultra thick, draw=darkgreen}
      [{\footnotesize\color{darkgreen}1}]
      [$x_4$, edge+={ultra thick, draw=darkgreen}
       [{\footnotesize\color{darkred}0}]
       [{\footnotesize\color{darkgreen}1}, edge+={ultra thick, draw=darkgreen}]
      ]
     ]
    ]
  ]
\end{forest}}
    \end{tabular}
\end{adjustbox}
\caption{Random Forest trained with Scikit-learn on Boolean 
		dataset \emph{xd6} using 4 trees and a max-depth of 4.}
\label{fig:ex1}
\end{figure}
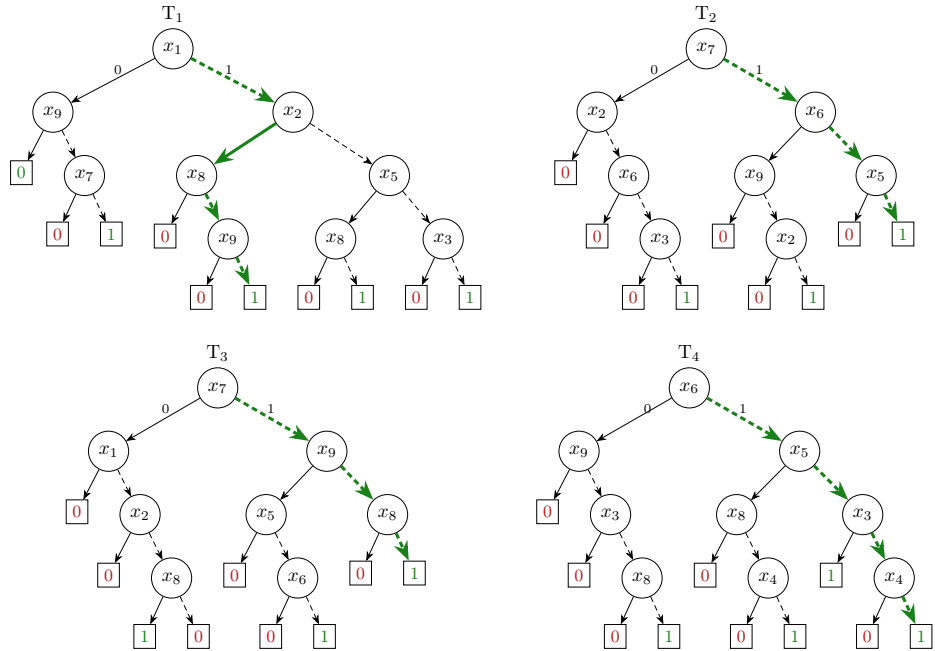

\begin{example}
Consider the simple example RF (using majority voting) shown in
\cref{fig:ex1}, which was obtained from Scikit-learn and trained on a
full binary dataset ``xd6''  (9 binary features) with 4 decision trees.
For the input $(1,0,1,1,1,1,1,1,1)$, all four trees predict 1 
(decision path is highlighted in each tree), so the prediction of this
RF is 1 in this case.
PyXAI returns an AXp $\{7, 8, 9\}$, meaning that for any point
$\mbf{x}$ such that $x_7=1,x_8=1$, and $x_9=1$, the RF predicts 1.
However, our checker returns a witness point
$(x_1=1,x_2=1,x_3=0,x_4=1,x_5=1,x_6=0,x_7=1,x_8=1,x_9=1)$ 
such that the first and fourth trees predict 0, while the second and
third trees predict 1. 
(The prediction is tied, and according to lexicographic tie-breaking
of majority vote, the RF should return 0.)
The RF classifier confirms that the returned witness point predicts 0,
which means that the AXp returned by PyXAI is incorrect.
\end{example}

\paragraph{Example of $\bm{\neg\wcxp}$ error in PyXAI.}
%
We describe another simple RF for which PyXAI computes an CXp
$\fml{Y}$, which is shown not to even be a WCXp. This corresponds to
case C2b in~\cref{tab:cxp-cases}; however, our analysis proves that
the explanation computed by PyXAI in this case is guaranteed to be
incorrect. (As noted earlier in the paper, for the cases C2b but also
A3a, and without the aid of manual analysis as used below, we require
the reference formal explainer $R$ and the second formal explainer $S$
to agree on their assessment, since in these cases we do not have a
witness to help in proving the result.)

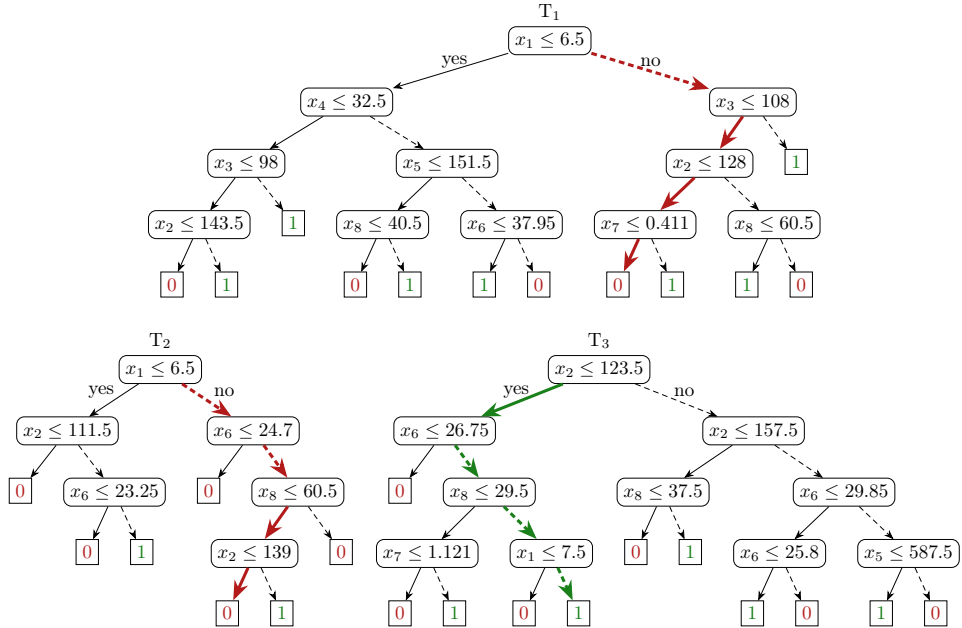
\begin{figure}[t]
  \centering
  \begin{adjustbox}{center}
    \setlength{\tabcolsep}{2pt}
    \def\arraystretch{3}
    \begin{tabular}{cc}
      \multicolumn{2}{c}{
       \scalebox{0.725}{\forestset{
  BDT/.style={
    for tree={
      l=1.125cm,s sep=0.575cm,
      if n children=0{}{rounded corners},
      draw,
      edge={
        my edge
      },
      if n=2 {
        edge+={0 my edge},
      }{},
    }
  },
}

\begin{forest}
  tikz+={\node[anchor=south] at (.north) {\text{T\(_1\)}};},
  BDT
  [$x_1 \leq 6.5$
    [$x_4 \leq 32.5$, edge label={node[near start,left,xshift=-2.75pt] {{ yes}}}
      [$x_3 \leq 98$
        [$x_2 \leq 143.5$
          [{\color{darkred}0}]
          [{\color{darkgreen}1}]
        ]
        [{\color{darkgreen}1}]
      ]
      [$x_5 \leq 151.5$
        [$x_8 \leq 40.5$
          [{\color{darkred}0}]
          [{\color{darkgreen}1}]
        ]
        [$x_6 \leq 37.95$
          [{\color{darkgreen}1}]        
          [{\color{darkred}0}]
        ]
      ]
    ]
    [$x_3 \leq 108$, edge label={node[near start,right,xshift=2.75pt] {{ no}}}, edge+={ultra thick, draw=darkred}
      [$x_2 \leq 128$, edge+={ultra thick, draw=darkred}
        [$x_7 \leq 0.411$, edge+={ultra thick, draw=darkred}
          [{\color{darkred}0}, edge+={ultra thick, draw=darkred}]
          [{\color{darkgreen}1}]
        ]
        [$x_8 \leq 60.5$
          [{\color{darkgreen}1}]      
          [{\color{darkred}0}]
        ] 
      ]
      [{\color{darkgreen}1}]
    ] 
 ]
\end{forest}} } \\
       \scalebox{0.725}{\forestset{
  BDT/.style={
    for tree={
      l=1.125cm,s sep=0.575cm,
      if n children=0{}{rounded corners},
      draw,
      edge={
        my edge
      },
      if n=2 {
        edge+={0 my edge},
      }{},
    }
  },
}

\begin{forest}
  tikz+={\node[anchor=south] at (.north) {\text{T\(_2\)}};},
  BDT
  [$x_1 \leq 6.5$
    [$x_2 \leq 111.5$,  edge label={node[near start,left,xshift=-2.75pt] {{ yes}}}
      [{\color{darkred}0}]
      [$x_6 \leq 23.25$
        [{\color{darkred}0}]
        [{\color{darkgreen}1}]
      ]
    ]
    [$x_6 \leq 24.7$, edge label={node[near start,right,xshift=2.75pt] {{ no}}}, edge+={ultra thick, draw=darkred}
      [{\color{darkred}0}]
      [$x_8 \leq 60.5$, edge+={ultra thick, draw=darkred}
        [$x_2 \leq 139$, edge+={ultra thick, draw=darkred}
          [{\color{darkred}0}, edge+={ultra thick, draw=darkred}]
          [{\color{darkgreen}1}]
        ]
        [{\color{darkred}0}] 
      ]
    ] 
 ]
\end{forest}} &
       \scalebox{0.725}{\forestset{
  BDT/.style={
    for tree={
      l=1.125cm,s sep=0.575cm,
      if n children=0{}{rounded corners},
      draw,
      edge={
        my edge
      },
      if n=2 {
        edge+={0 my edge},
      }{},
    }
  },
}

\begin{forest}
  tikz+={\node[anchor=south] at (.north) {\text{T\(_3\)}};},
  BDT
  [$x_2 \leq 123.5$
    [$x_6 \leq 26.75$, edge label={node[near start,left,xshift=-2.75pt] {{ yes}}}, edge+={ultra thick, draw=darkgreen}
      [{\color{darkred}0}]
      [$x_8 \leq 29.5$, edge+={ultra thick, draw=darkgreen}
         [$x_7 \leq 1.121$
            [{\color{darkred}0}]
            [{\color{darkgreen}1}]         
         ]
         [$x_1 \leq  7.5$, edge+={ultra thick, draw=darkgreen}
            [{\color{darkred}0}]
            [{\color{darkgreen}1}, edge+={ultra thick, draw=darkgreen}]            
         ]
      ]
    ]
    [$x_2 \leq 157.5$,  edge label={node[near start,right,xshift=2.75pt] {{ no}}}
      [$x_8 \leq 37.5$
          [{\color{darkred}0}]
          [{\color{darkgreen}1}]
      ]    
      [$x_6 \leq 29.85$
        [$x_6 \leq 25.8$
          [{\color{darkgreen}1}]
          [{\color{darkred}0}]
        ]
        [$x_5 \leq 587.5$
          [{\color{darkgreen}1}]
          [{\color{darkred}0}] 
        ] 
      ]
    ] 
 ]
\end{forest}}
    \end{tabular}
  \end{adjustbox}
  \caption{Random Forest trained with Scikit-learn on the \emph{Pima} dataset (3 trees, max-depth of 3).}
  \label{fig:ex2}
\end{figure}

\begin{example}
Consider the simple example RF (majority vote) shown in
\cref{fig:ex2},  trained on the ``Pima'' dataset (which has 8
real-valued features) using Scikit-learn and the tree number fixed to 3.
For the input $\mbf{v}=(9.0,57.0,\allowbreak80.0,37.0,\allowbreak0.0,32.8,0.096,41.0)$,
trees T$_1$ and T$_2$ predict 0 while the third tree predicts 1. 
(The decision path of each trees is highlighted in \cref{fig:ex2}.) 
As a result, the prediction of the RF is class 0 by majority vote.
We ran PyXAI on $\mbf{v}$ to generate a CXp $\{1\}$.
This explanation suggests that  
if we fix $x_2=57.0, x_3=80.0, x_4=37.0, x_5=0.0, x_6=32.8, x_7=0.096, x_8=41.0$,
and allow $x_1$ to take value other than $9.0$,
then RF's prediction changes to 1.
However, it can be verified by inspection that $\{0\}$ is not a correct CXp.
First, observe that $T_2$ always predicts 0 regardless of the value of $x_1$.
Second, if $x_1$ takes a value such that $T_3$ predicts 1, then this value must satisfy $x_1 > 7.5$.
Since $x_1 > 7.5$ implies $x_1 > 6.5$, $T_1$ predicts 0;
in this case, the RF predicts 0.
Conversely, if $x_1$ takes a value such that $T_3$ predicts 0, then the RF also predicts 0.
Hence, no prediction flip occurs, and the CXp is incorrect.
\end{example}


\section{Conclusions and Future Work} \label{sec:conc}

Given a machine learning model $M$ and an instance $(\mbf{v},c)$ to
explain, this paper proposes a framework for validating the
explanations obtained with an untrusted target formal explainer $T$,
using queries to other possibly untrusted explainers $R$ and $S$, but
also queries to the machine learning model $M$.
The paper also proposes a framework for validating the results
obtained with the explainer $R$.
The experimental setting considered a publicly available formal
explainer PyXAI~\cite{marquis-ijcai24a} as $T$,
RFxpl~\cite{ims-ijcai21} as $R$ and XReason~\cite{iisms-aaai22} as
$S$.

The experiments reveal the existence of bug(s) in the recently
published formal explanation toolkit PyXAI~\cite{marquis-ijcai24a}. 
As shown by the experiments, the percentage of incorrectly computed
explanations is significant, being observed in most of the test cases
considered.
In addition, the analysis of two concrete case studies sheds light on
some of PyXAI's bugs.
%
Since the detected bugs occur in in both AXp and CXp 
explanations --- either in
terms of correctness or in terms of redundacy --- one may anticipate that 
the bugs originate from different sources. 
Furthermore, the uncovered bugs in PyXAI raise not only a concern
about the general reliability of formal explainers, but also
conclusions that have been drawn with PyXAI.
Future research will extend this paper to consider other tools, other
families of classifiers, but also regression models.
%

\section*{Acknowledgments}
This work was supported in part by the Spanish Government under
grant PID 2023-152814OB-I00, and by ICREA starting funds.
The last author (JMS) also acknowledges the extra incentive provided
by the ERC in not funding this research.
%
\bibliographystyle{splncs04}
%

\newtoggle{mkbbl}

\settoggle{mkbbl}{false}

\iftoggle{mkbbl}{
    \bibliography{refs,xtra}

\begin{thebibliography}{10}
\providecommand{\url}[1]{\texttt{#1}}
\providecommand{\urlprefix}{URL }
\providecommand{\doi}[1]{https://doi.org/#1}

\bibitem{amgoud-ijcai22}
Amgoud, L., Ben{-}Naim, J.: Axiomatic foundations of explainability. In: IJCAI.
  pp. 636--642 (2022)

\bibitem{katz-ecai24}
Amir, G., Bassan, S., Katz, G.: Hard to explain: On the computational hardness
  of in-distribution model interpretation. In: ECAI. pp. 818--825 (2024)

\bibitem{barcelo-nips22}
Arenas, M., Barcel{\'{o}}, P., Orth, M.A.R., Subercaseaux, B.: On computing
  probabilistic explanations for decision trees. In: NeurIPS (2022)

\bibitem{asuncion2007uci}
Asuncion, A., Newman, D., et~al.: Uci machine learning repository.
  \url{http://archive.ics.uci.edu/ml} (2007)

\bibitem{marquis-kr21}
Audemard, G., Bellart, S., Bounia, L., Koriche, F., Lagniez, J., Marquis, P.:
  On the computational intelligibility of boolean classifiers. In: KR. pp.
  74--86 (2021)

\bibitem{marquis-ijcai22b}
Audemard, G., Bellart, S., Bounia, L., Koriche, F., Lagniez, J., Marquis, P.:
  On preferred abductive explanations for decision trees and random forests.
  In: IJCAI. pp. 643--650 (2022)

\bibitem{marquis-dke22}
Audemard, G., Bellart, S., Bounia, L., Koriche, F., Lagniez, J., Marquis, P.:
  On the explanatory power of boolean decision trees. Data Knowl. Eng.
  \textbf{142},  102088 (2022)

\bibitem{marquis-aaai22}
Audemard, G., Bellart, S., Bounia, L., Koriche, F., Lagniez, J., Marquis, P.:
  Trading complexity for sparsity in random forest explanations. In: AAAI. pp.
  5461--5469 (2022)

\bibitem{marquis-ijcai23a}
Audemard, G., Bellart, S., Lagniez, J., Marquis, P.: Computing abductive
  explanations for boosted regression trees. In: IJCAI. pp. 3432--3441 (2023)

\bibitem{marquis-aistats23}
Audemard, G., Lagniez, J., Marquis, P., Szczepanski, N.: Computing abductive
  explanations for boosted trees. In: AISTATS. pp. 4699--4711 (2023)

\bibitem{marquis-ecai23}
Audemard, G., Lagniez, J., Marquis, P., Szczepanski, N.: On contrastive
  explanations for tree-based classifiers. In: ECAI. pp. 117--124 (2023)

\bibitem{marquis-ijcai24a}
Audemard, G., Lagniez, J., Marquis, P., Szczepanski, N.: {PyXAI}: An {XAI}
  library for tree-based models. In: IJCAI. pp. 8601--8605 (2024)

\bibitem{barcelo-pods25}
Barcel{\'{o}}, P., Kozachinskiy, A., Romero, M., Subercaseaux, B., Verschae,
  J.: Explaining \emph{k}-nearest neighbors: Abductive and counterfactual
  explanations. Proc. {ACM} Manag. Data  \textbf{3}(2),  97:1--97:26 (2025)

\bibitem{barcelo-nips20}
Barcel{\'{o}}, P., Monet, M., P{\'{e}}rez, J., Subercaseaux, B.: Model
  interpretability through the lens of computational complexity. In: NeurIPS
  (2020)

\bibitem{katz-icml24}
Bassan, S., Amir, G., Katz, G.: Local vs. global interpretability: {A}
  computational complexity perspective. In: ICML (2024)

\bibitem{katz-tacas23}
Bassan, S., Katz, G.: Towards formal {XAI:} formally approximate minimal
  explanations of neural networks. In: TACAS. pp. 187--207 (2023)

\bibitem{bertot-bk04}
Bertot, Y., Cast{\'{e}}ran, P.: Interactive Theorem Proving and Program
  Development - Coq'Art: The Calculus of Inductive Constructions. Texts in
  Theoretical Computer Science. An {EATCS} Series, Springer (2004)

\bibitem{sat-handbook21}
Biere, A., Heule, M., van Maaren, H., Walsh, T. (eds.): Handbook of
  Satisfiability - Second Edition, Frontiers in Artificial Intelligence and
  Applications, vol.~336. {IOS} Press (2021)

\bibitem{mazure-cikm21}
Boumazouza, R., Alili, F.C., Mazure, B., Tabia, K.: {ASTERYX:} {A}
  model-agnostic sat-based approach for symbolic and score-based explanations.
  In: CIKM. pp. 120--129 (2021)

\bibitem{bounia-uai25}
Bounia, L.: Using submodular optimization to approximate minimum-size abductive
  path explanations for tree-based models. In: Chiappa, S., Magliacane, S.
  (eds.) UAI. pp. 388--397 (2025)

\bibitem{breiman-ml01}
Breiman, L.: Random forests. Mach. Learn.  \textbf{45}(1),  5--32 (2001)

\bibitem{ccms-kr23}
Carbonnel, C., Cooper, M.C., Marques{-}Silva, J.: Tractable explaining of
  multivariate decision trees. In: KR. pp. 127--135 (2023)

\bibitem{guestrin-kdd16b}
Chen, T., Guestrin, C.: {XGBoost}: {A} scalable tree boosting system. In: KDD.
  pp. 785--794 (2016)

\bibitem{chen2016xgboost}
Chen, T., Guestrin, C.: Xgboost: A scalable tree boosting system. In:
  Proceedings of the 22nd acm sigkdd international conference on knowledge
  discovery and data mining. pp. 785--794 (2016)

\bibitem{cms-cp21}
Cooper, M.C., Marques{-}Silva, J.: On the tractability of explaining decisions
  of classifiers. In: CP. pp. 21:1--21:18 (2021)

\bibitem{cms-aij23}
Cooper, M.C., Marques{-}Silva, J.: Tractability of explaining classifier
  decisions. Artif. Intell.  \textbf{316},  103841 (2023)

\bibitem{darwiche-lics23}
Darwiche, A.: Logic for explainable {AI}. In: {LICS}. pp. 1--11 (2023)

\bibitem{darwiche-ecai20}
Darwiche, A., Hirth, A.: On the reasons behind decisions. In: ECAI. pp.
  712--720 (2020)

\bibitem{darwiche-jlli23}
Darwiche, A., Hirth, A.: On the (complete) reasons behind decisions. J. Log.
  Lang. Inf.  \textbf{32}(1),  63--88 (2023)

\bibitem{darwiche-aaai22}
Darwiche, A., Ji, C.: On the computation of necessary and sufficient
  explanations. In: AAAI. pp. 5582--5591 (2022)

\bibitem{katz-tap25}
Desmartin, R., Isac, O., Passmore, G.O., Komendantskaya, E., Stark, K., Katz,
  G.: A certified proof checker for deep neural network verification in
  {I}mandra. In: TAP. pp. 1:1--1:21 (2025)

\bibitem{rubin-aaai22}
Gorji, N., Rubin, S.: Sufficient reasons for classifier decisions in the
  presence of domain constraints. In: AAAI. pp. 5660--5667 (2022)

\bibitem{gunning-darpa19}
Gunning, D., Aha, D.W.: {DARPA's} explainable artificial intelligence {(XAI)}
  program. {AI} Mag.  \textbf{40}(2),  44--58 (2019)

\bibitem{gunning-sr19}
Gunning, D., Stefik, M., Choi, J., Miller, T., Stumpf, S., Yang, G.: {XAI} -
  explainable artificial intelligence. Sci. Robotics  \textbf{4}(37) (2019)

\bibitem{han2022adbench}
Han, S., Hu, X., Huang, H., Jiang, M., Zhao, Y.: Adbench: Anomaly detection
  benchmark. Advances in neural information processing systems  \textbf{35},
  32142--32159 (2022)

\bibitem{hcmpms-tacas23}
Huang, X., Cooper, M.C., Morgado, A., Planes, J., Marques{-}Silva, J.: Feature
  necessity {\&} relevancy in {ML} classifier explanations. In: TACAS. pp.
  167--186 (2023)

\bibitem{hiicams-aaai22}
Huang, X., Izza, Y., Ignatiev, A., Cooper, M.C., Asher, N., Marques{-}Silva,
  J.: Tractable explanations for {d-DNNF} classifiers. In: AAAI. pp. 5719--5728
  (2022)

\bibitem{hiims-kr21}
Huang, X., Izza, Y., Ignatiev, A., Marques{-}Silva, J.: On efficiently
  explaining graph-based classifiers. In: KR. pp. 356--367 (2021)

\bibitem{hims-aaai23}
Huang, X., Izza, Y., Marques{-}Silva, J.: Solving explainability queries with
  quantification: The case of feature relevancy. In: AAAI. pp. 3996--4006
  (2023)

\bibitem{hms-ecai23}
Huang, X., Marques{-}Silva, J.: From decision trees to explained decision sets.
  In: ECAI. pp. 1100--1108 (2023)

\bibitem{hms-tap23}
Hurault, A., Marques{-}Silva, J.: Certified logic-based explainable {AI} - the
  case of monotonic classifiers. In: TAP. pp. 51--67 (2023)

\bibitem{ignatiev-ijcai20}
Ignatiev, A.: Towards trustable explainable {AI}. In: IJCAI. pp. 5154--5158
  (2020)

\bibitem{iisms-aaai22}
Ignatiev, A., Izza, Y., Stuckey, P.J., Marques{-}Silva, J.: Using maxsat for
  efficient explanations of tree ensembles. In: AAAI. pp. 3776--3785 (2022)

\bibitem{ims-sat21}
Ignatiev, A., Marques{-}Silva, J.: {SAT}-based rigorous explanations for
  decision lists. In: SAT. pp. 251--269 (2021)

\bibitem{inams-aiia20}
Ignatiev, A., Narodytska, N., Asher, N., Marques{-}Silva, J.: From contrastive
  to abductive explanations and back again. In: AIxIA. pp. 335--355 (2020)

\bibitem{inms-aaai19}
Ignatiev, A., Narodytska, N., Marques{-}Silva, J.: Abduction-based explanations
  for machine learning models. In: AAAI. pp. 1511--1519 (2019)

\bibitem{inms-nips19}
Ignatiev, A., Narodytska, N., Marques{-}Silva, J.: On relating explanations and
  adversarial examples. In: NeurIPS. pp. 15857--15867 (2019)

\bibitem{ihincms-ijar23}
Izza, Y., Huang, X., Ignatiev, A., Narodytska, N., Cooper, M.C.,
  Marques{-}Silva, J.: On computing probabilistic abductive explanations. Int.
  J. Approx. Reason.  \textbf{159},  108939 (2023)

\bibitem{ihmpims-kr24}
Izza, Y., Huang, X., Morgado, A., Planes, J., Ignatiev, A., Marques{-}Silva,
  J.: Distance-restricted explanations: Theoretical underpinnings {\&}
  efficient implementation. In: KR (2024)

\bibitem{iirmss-ijcai25}
Izza, Y., Ignatiev, A., Rubin, S., Marques-Silva, J., Stuckey, P.J.: Most
  general explanations of tree ensembles. In: IJCAI (2025)

\bibitem{iisms-aaai24}
Izza, Y., Ignatiev, A., Stuckey, P.J., Marques{-}Silva, J.: Delivering inflated
  explanations. In: AAAI. pp. 12744--12753 (2024)

\bibitem{ims-ijcai21}
Izza, Y., Marques{-}Silva, J.: On explaining random forests with {SAT}. In:
  IJCAI. pp. 2584--2591 (2021)

\bibitem{imms-ecai24}
Izza, Y., Meel, K.S., Marques{-}Silva, J.: Locally-minimal probabilistic
  explanations. In: ECAI. pp. 1092--1099 (2024)

\bibitem{darwiche-jelia23}
Ji, C., Darwiche, A.: A new class of explanations for classifiers with
  non-binary features. In: JELIA. pp. 106--122 (2023)

\bibitem{lhms-aaai25}
L{\'{e}}toff{\'{e}}, O., Huang, X., Marques{-}Silva, J.: Towards trustable
  {SHAP} scores. In: AAAI. pp. 18198--18208 (2025)

\bibitem{lorini-clar21}
Liu, X., Lorini, E.: A logic for binary classifiers and their explanation. In:
  CLAR. pp. 302--321 (2021)

\bibitem{lorini-wollic22}
Liu, X., Lorini, E.: A logic of "black box" classifier systems. In: WoLLIC. pp.
  158--174 (2022)

\bibitem{lorini-jlc23}
Liu, X., Lorini, E.: A unified logical framework for explanations in classifier
  systems. J. Log. Comput.  \textbf{33}(2),  485--515 (2023)

\bibitem{lundberg-nips17}
Lundberg, S.M., Lee, S.: A unified approach to interpreting model predictions.
  In: NeurIPS. pp. 4765--4774 (2017)

\bibitem{kwiatkowska-ijcai21a}
Malfa, E.L., Michelmore, R., Zbrzezny, A.M., Paoletti, N., Kwiatkowska, M.: On
  guaranteed optimal robust explanations for {NLP} models. In: IJCAI. pp.
  2658--2665 (2021)

\bibitem{ms-rw22}
Marques{-}Silva, J.: Logic-based explainability in machine learning. In:
  Reasoning Web. pp. 24--104 (2022)

\bibitem{ms-iceccs23}
Marques{-}Silva, J.: Disproving {XAI} myths with formal methods - initial
  results. In: ICECCS. pp. 12--21 (2023)

\bibitem{ms-isola24}
Marques{-}Silva, J.: Logic-based explainability: Past, present and future. In:
  ISoLA. pp. 181--204 (2024)

\bibitem{msgcin-nips20}
Marques{-}Silva, J., Gerspacher, T., Cooper, M.C., Ignatiev, A., Narodytska,
  N.: Explaining naive bayes and other linear classifiers with polynomial time
  and delay. In: NeurIPS (2020)

\bibitem{msgcin-icml21}
Marques{-}Silva, J., Gerspacher, T., Cooper, M.C., Ignatiev, A., Narodytska,
  N.: Explanations for monotonic classifiers. In: ICML. pp. 7469--7479 (2021)

\bibitem{msh-cacm24}
Marques{-}Silva, J., Huang, X.: Explainability is \emph{Not} a game. Commun.
  {ACM}  \textbf{67}(7),  66--75 (2024)

\bibitem{msi-frai23}
Marques{-}Silva, J., Ignatiev, A.: No silver bullet: interpretable {ML} models
  must be explained. Frontiers Artif. Intell.  \textbf{6} (2023)

\bibitem{msllm-ijcai25}
Marques-Silva, J., Lefebre-Lobaina, J., Martinez, V.: Efficient and rigorous
  model-agnostic explanations. In: IJCAI (2025)

\bibitem{msm-ijcai20}
Marques{-}Silva, J., Menc{\'{\i}}a, C.: Reasoning about inconsistent formulas.
  In: IJCAI. pp. 4899--4906 (2020)

\bibitem{katz-aistats25}
Marzouk, R., Bassan, S., Katz, G., de~la Higuera, C.: On the computational
  tractability of the (many) shapley values. In: AISTATS. pp. 3691--3699 (2025)

\bibitem{moura-cade21}
de~Moura, L., Ullrich, S.: The lean 4 theorem prover and programming language.
  In: CADE. pp. 625--635 (2021)

\bibitem{olson2017pmlb}
Olson, R.S., La~Cava, W., Orzechowski, P., Urbanowicz, R.J., Moore, J.H.: Pmlb:
  a large benchmark suite for machine learning evaluation and comparison.
  BioData mining  \textbf{10},  1--13 (2017)

\bibitem{scikitlearn}
Pedregosa, F., et~al.: Scikit-learn: Machine learning in {P}ython. Journal of
  Machine Learning Research  \textbf{12},  2825--2830 (2011)

\bibitem{pedregosa2011scikit}
Pedregosa, F., Varoquaux, G., Gramfort, A., Michel, V., Thirion, B., Grisel,
  O., Blondel, M., Prettenhofer, P., Weiss, R., Dubourg, V., et~al.:
  Scikit-learn: Machine learning in python. the Journal of machine Learning
  research  \textbf{12},  2825--2830 (2011)

\bibitem{guestrin-kdd16}
Ribeiro, M.T., Singh, S., Guestrin, C.: "why should {I} trust you?": Explaining
  the predictions of any classifier. In: KDD. pp. 1135--1144 (2016)

\bibitem{guestrin-aaai18}
Ribeiro, M.T., Singh, S., Guestrin, C.: Anchors: High-precision model-agnostic
  explanations. In: AAAI. pp. 1527--1535 (2018)

\bibitem{seshia-cacm22}
Seshia, S.A., Sadigh, D., Sastry, S.S.: Toward verified artificial
  intelligence. Commun. {ACM}  \textbf{65}(7),  46--55 (2022)

\bibitem{darwiche-ijcai18}
Shih, A., Choi, A., Darwiche, A.: A symbolic approach to explaining bayesian
  network classifiers. In: IJCAI. pp. 5103--5111 (2018)

\bibitem{kutyniok-jair21}
W{\"{a}}ldchen, S., MacDonald, J., Hauch, S., Kutyniok, G.: The computational
  complexity of understanding binary classifier decisions. J. Artif. Intell.
  Res.  \textbf{70},  351--387 (2021)

\bibitem{wetzler2014drat}
Wetzler, N., Heule, M.J., Hunt~Jr, W.A.: Drat-trim: Efficient checking and
  trimming using expressive clausal proofs. In: International Conference on
  Theory and Applications of Satisfiability Testing. pp. 422--429. Springer
  (2014)

\bibitem{barrett-nips23}
Wu, M., Wu, H., Barrett, C.W.: {VeriX}: Towards verified explainability of deep
  neural networks. In: NeurIPS (2023)

\bibitem{yisnms-aaai23}
Yu, J., Ignatiev, A., Stuckey, P.J., Narodytska, N., Marques{-}Silva, J.:
  Eliminating the impossible, whatever remains must be true: On extracting and
  applying background knowledge in the context of formal explanations. In:
  AAAI. pp. 4123--4131 (2023)

\bibitem{zhou-bk12}
Zhou, Z.H.: Ensemble methods: foundations and algorithms. Chapman and Hall/CRC
  (2012)

\bibitem{zhou-bk21}
Zhou, Z.: Machine Learning. Springer (2021)

\end{thebibliography}
}{

}

\clearpage

\appendix
\section*{Appendices}

\crefname{subappendix}{section}{sections}

\section{Results for Boosted Trees} \label[appendix]{sec:bts}

\begin{table}[ht]
  \centering
  \renewcommand{\tabcolsep}{0.725em}
\begin{tabular}{lrrr}
\toprule[1.2pt]
Dataset             &  \%$\:[\neg\msf{WCXp}]$ & \%$\:[\msf{WCXp} \land \neg \msf{CXp}]$ & \%$\:[\msf{CXp}]$ \\
\midrule[1.2pt]
12\_fault           & 1.5\%        & 51.0\%            & 47.5\%     \\
21\_Lymphography    & 0.0\%        & 5.4\%             & 94.6\%     \\
29\_Pima            & 1.0\%        & 59.0\%            & 40.0\%     \\
30\_satellite       & 3.0\%        & 73.0\%            & 24.0\%     \\
31\_satimage-2      & 0.0\%        & 99.5\%            & 0.5\%      \\
33\_skin            & 0.0\%        & 62.4\%            & 37.6\%     \\
37\_Stamps          & 0.0\%        & 77.0\%            & 23.0\%     \\
4\_breastw          & 2.0\%        & 75.2\%            & 22.8\%     \\
6\_cardio           & 0.0\%        & 87.0\%            & 13.0\%     \\
7\_Cardiotocography & 0.5\%        & 46.5\%            & 53.0\%     \\
appendicitis        & 0.0\%        & 53.8\%            & 46.2\%     \\
banknote            & 0.0\%        & 64.5\%            & 35.5\%     \\
biodegradation      & 1.5\%        & 61.0\%            & 37.5\%     \\
glass2              & 0.0\%        & 42.6\%            & 57.4\%     \\
heart-c             & 0.5\%        & 14.5\%            & 85.0\%     \\
ionosphere          & 0.0\%        & 44.0\%            & 56.0\%     \\
magic               & 0.5\%        & 68.5\%            & 31.0\%     \\
mofn-3-7-10         & 0.0\%        & 0.0\%             & 100.0\%    \\
phoneme             & 0.5\%        & 63.5\%            & 36.0\%     \\
ring                & 0.0\%        & 97.5\%            & 2.5\%      \\
sonar               & 0.0\%        & 72.5\%            & 27.5\%     \\
spambase            & 0.5\%        & 81.0\%            & 18.5\%     \\
spectf              & 0.0\%        & 72.5\%            & 27.5\%     \\
twonorm             & 0.0\%        & 98.0\%            & 2.0\%      \\
wdbc                & 0.0\%        & 92.0\%            & 8.0\%      \\
wpbc                & 0.0\%        & 77.3\%            & 22.7\%     \\
xd6                 & 0.0\%        & 0.0\%             & 100.0\%   \\
\bottomrule[1.2pt]
\end{tabular}
\medskip
\caption{
    Assessing the validity of CXps for BTs obtained using the PyXAI
    tool with default settings (using the SCIP MIP solver).
    \%$[\msf{CXp}]$ denotes the fraction of computed CXps that are valid,
    \%$[\neg\msf{WCXp}]$ denotes the fraction of computed CXps that incorrect,
    and \%$[\msf{WCXp}\land\neg\msf{CXp}]$ denotes the fraction of computed CXps that are redundant.
    All BTs have 50 trees per class. The number of instances is 200, or 
    all instances if there are fewer than 200; no timeout was observed.
}
\label{tab:bt_pyxai_cxp_quality}
\end{table}

\paragraph{Validation of PyXAI's CXps for BTs.}
\cref{tab:bt_pyxai_cxp_quality} reports a summary of results of assessing 
PyXAI's contrastive explanations for boost trees (BTs). In this assessment, the reference 
explainer $R$ is XReason (leveraging an SMT reasoner\footnote{%
Note that XReason also includes a MaxSAT-based reasoner, which applies 
rounding to prediction scores and operates with decimal approximations, whereas 
the SMT-based approach uses exact real-valued scores. Therefore, we deliberately 
employ SMT-based approach for precise reasoning over the BT classifiers.}) and 
the target explainer $T$ is PyXAI (leveraging a MIP reasoner).  
As SMT oracle does not implement a proof trace $\prooftrace_R(\fml{X},c)$, we do not 
compute a proof of $R$ in steps C2 and C4.
As can be observed from \cref{tab:bt_pyxai_cxp_quality}, XReason reported several 
datasets for which the CXps returned by PyXAI are incorrect (10 out of 27), 
as well as redundant CXps for most benchmarks (except for 2 binary datasets where 
the CXps are found to be correct). These results are consistent with our observations 
for RFs, suggesting that the PyXAI explainer for BTs is also affected by bugs. 
However, the underlying causes may differ, since in this case there are no majority-vote 
ties or discrepancies related to fully binary data.

Besides, we emphasize that we do not report any assessments of AXps for BTs 
as PyXAI returns an exception at the first line of the invoked method.

\paragraph{Example of $\bm{\neg\wcxp}$ error in PyXAI.}
We describe a simple BT for which PyXAI computes an CXp
$\fml{X}$, which is shown not to be a WCXp.

\begin{figure}[ht]
  \centering
  \begin{adjustbox}{center}
    \setlength{\tabcolsep}{8pt}
    \def\arraystretch{3}
    \begin{tabular}{c}
       \scalebox{0.55}{\forestset{
  BDT/.style={
    for tree={
      l=1.125cm,s sep=0.575cm,
      if n children=0{}{rounded corners},
      draw,
      edge={
        my edge
      },
      if n=2 {
        edge+={0 my edge},
      }{},
    }
  },
}

\begin{forest}
  tikz+={\node[anchor=south] at (.north) {\text{T\(_1\)}};},
  BDT
[$x_4 < 0.631000042$
  [$x_4 < -0.308499992$, edge label={node[near start,left,xshift=-2.75pt] {{ yes}}}
    [$x_2 < 0.817999959$
      [$0.0290748924$]
      [$-0.33644861$]
    ]
    [$x_3 < 1.449$
      [$-0.541057169$]
      [$0.120000005$]
    ]
  ]
  [$x_1 < 1.403$, edge label={node[near start,right,xshift=2.75pt] {{ no}}}, edge+={ultra thick, draw=darkred}
    [$x_1 < 0.277500004$
      [$-0.0100278556$]
      [$0.226630449$]
    ]
    [$x_3 < 1.02250004$, edge+={ultra thick, draw=darkred}
      [$-0.564179122$, edge+={ultra thick, draw=darkred}]
      [$-0.0461538509$]
    ]
  ]
]
\end{forest}} \\
       \scalebox{0.55}{\forestset{
  BDT/.style={
    for tree={
      l=1.125cm,s sep=0.575cm,
      if n children=0{}{rounded corners},
      draw,
      edge={
        my edge
      },
      if n=2 {
        edge+={0 my edge},
      }{},
    }
  },
}

\begin{forest}
  tikz+={\node[anchor=south] at (.north) {\text{T\(_2\)}};},
  BDT
[$x_4<0.702499986$
  [$x_3 < 0.685000002$,  edge label={node[near start,left,xshift=-2.75pt] {{ yes}}}, edge+={ultra thick, draw=darkred}
    [$x_1<0.172499999$, edge+={ultra thick, draw=darkred}
      [$-0.153566822$]
      [$-0.456195921$, edge+={ultra thick, draw=darkred}]
    ]
    [$x_5<0.621999979$
      [$-0.241189703$]
      [$0.352639019$]
    ]
  ]
  [$x_1<1.324$, edge label={node[near start,right,xshift=2.75pt] {{ no}}}
    [$x_1<0.159500003$
      [$-0.0891945809$]
      [$0.154747352$]
    ]
    [$x_3<1.31799996$
      [$-0.421002328$]
      [$0.0654052421$]
    ]
  ]
]
\end{forest}}
    \end{tabular}
  \end{adjustbox}
  \caption{Boosted Tree trained with XGBoost on the \emph{phoneme} dataset (2 trees, max-depth of 2).}
  \label{fig:ex3}
\end{figure}
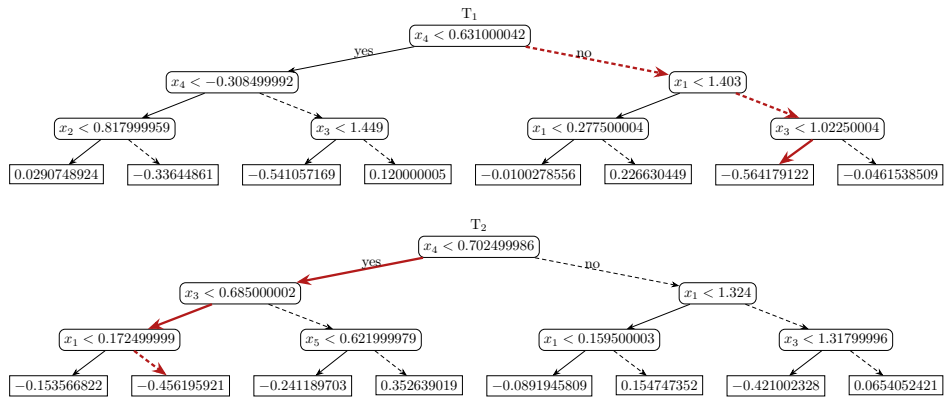

\begin{example}
Consider the toy example BT shown in \cref{fig:ex3} , trained on 
``Phoneme'' dataset (5 real-valued features) with 2 trees using 
XGBoost. 
For the input $\mbf{v}=(3.306,0.653,0.313,0.669,-0.218)$, trees 
$T_1$ returns score $-0.564179122$, $T_2$ returns $-0.456195921$, and 
aggregating the scores gives: $-1.020375043 < 0$. As a result 
the predicted class of this BT is 0 (i.e. negative score, $\kappa(\mbf{v}) < 0$).
We ran PyXAI on $\mbf{v}$ to generate a CXp $\{1\}$,
and to change the prediction of this BT, the aggregated score must be positive.
We observe that tree $T_2$'s prediction remains unchanged as it always return a negative score.
To change the prediction of this BT, tree $T_1$ must return a positive score.
Note that in $T_1$, if $x_1$ takes a value other than $3.306$,
the largest possible (positive) score for $T_1$ is $0.226630449$
(among $\{-0.564179122, -0.0100278556, 0.226630449\}$).
In this case, such a value must satisfy $0.277500004 \le x_1 < 1.403$.
This, in turn, implies $x_1 > 0.172499999$, which forces $T_2$ to predict $-0.456195921$;
so the final BT score remains negative.
Hence, no prediction flip occurs, and the explanation is incorrect.
\end{example}

\section{Training of RFs/BTs} \label[appendix]{sec:training}
For each dataset, a RF/BT classifier was trained using 80\% of the data for training and 20\% for testing.
All feature data types are numerical.

Regarding random forests, we utilize scikit-learn~\cite{pedregosa2011scikit} to learn a collection
of 50 decision trees with a maximum depth of 4. The individual tree classifiers are fed to the ensemble voting classifier 
of scikit-learn to produce majority vote predictions, rather than the probability-weighted voting 
used in scikit-learn’s default RF implementation.
The resulting trained models achieved a test accuracy of at least 74\%.

Regarding boosted trees, we utilize XGBoost~\cite{chen2016xgboost} to learn boost tree classifiers with 50 trees per class.
The resulting trained models also achieved a test accuracy of at least 74\%.

Both tools PyXAI and RFxpl are supporting scikit-learn format. Moreover, PyXAI provides 
its own tree ensemble class, which loads a scikit-learn model and overrides the prediction 
function to enforce majority voting for RFs.
To ensure consistency between the models used by RFxpl and PyXAI during verification, we apply 
the prediction function from PyXAI’s source implementation when checking witnesses.    
When reasoning about RFs with XReason, we use as input the SMT encoding 
generated by RFxpl, since XReason does not support the scikit-learn tree ensemble 
format and is designed exclusively for the XGBoost format.

\end{document}